\documentclass[11pt]{article}
\usepackage[final]{acl}
\usepackage{times}
\usepackage{latexsym}
\usepackage{enumitem}
\usepackage{soul}
\usepackage{xcolor}
\usepackage{makecell}
\usepackage{float}
\restylefloat{table}
\usepackage{placeins}
\usepackage{afterpage}
\usepackage[T1]{fontenc}
\usepackage[utf8]{inputenc}
\usepackage{microtype}
\usepackage{inconsolata}
\usepackage{graphicx}

\usepackage{tipa}
\usepackage{multirow}
\usepackage{amssymb}
\usepackage{tcolorbox}
\usepackage{geometry}
\usepackage{booktabs,tabularx}

\usepackage{bm}
\usepackage{amsmath}
\usepackage{colortbl}
\usepackage{siunitx}
\usepackage{pifont}
\newcommand{\cmark}{\ding{51}}%
\newcommand{\xmark}{\ding{55}}%
\usepackage{adjustbox}
\usepackage{xcolor}

\newcommand{\OURS}[1]{\textbf{SpidR-Adapt}}

\title{SpidR-Adapt: A Universal Speech Representation Model \\ for Few-Shot Adaptation}

\author{
 \textbf{Mahi Luthra\textsuperscript{*,1}},
 \textbf{Jiayi Shen\textsuperscript{*,1}},
 \textbf{Maxime Poli\textsuperscript{*,2}},
 \textbf{Angelo Ortiz Tandazo\textsuperscript{2}},
  \\
 \textbf{Yosuke Higuchi\textsuperscript{1}},
 \textbf{Youssef Benchekroun\textsuperscript{1}},
  \textbf{Martin Gleize\textsuperscript{1}},
 \textbf{Charles-Eric Saint-James\textsuperscript{1}},
 \\
 \textbf{Dongyan Lin\textsuperscript{1}},
 \textbf{Phillip Rust\textsuperscript{1}},
 \textbf{Angel Villar\textsuperscript{1}},
 \textbf{Surya Parimi\textsuperscript{1}},
 \textbf{Vanessa Stark\textsuperscript{1}},
 \textbf{Rashel Moritz\textsuperscript{1}},
  \\
 \textbf{Juan Pino\textsuperscript{1}},
 \textbf{Yann LeCun\textsuperscript{1}},
  \textbf{Emmanuel Dupoux\textsuperscript{1,2}}
\\
\\
 \textsuperscript{*} Equal contribution,
 \textsuperscript{1} Meta AI,
 \textsuperscript{2} ENS-PSL, EHESS, CNRS
\\
 \small{
   \textbf{Correspondence:} \href{jiayishen@meta.com}{jiayishen@meta.com, mahiluthra@meta.com}
 }
}

\begin{document}
\maketitle

\begin{abstract}
Human infants, with only a few hundred hours of speech exposure, acquire basic units of new languages, highlighting a striking efficiency gap compared to the data-hungry self-supervised speech models. To address this gap, this paper introduces \textbf{SpidR-Adapt} for rapid adaptation of speech units to new languages using minimal unlabeled data. We cast such low-resource speech representation learning as a meta-learning problem and construct a multi-task adaptive pre-training (MAdaPT) protocol which formulates the adaptation process as a bi-level optimization framework. To enable scalable meta-training under this framework, we propose a novel heuristic solution, first-order bi-level optimization (FOBLO), avoiding heavy computation costs.  Finally, we stabilize meta-training by using a robust initialization through interleaved supervision which alternates self-supervised and supervised objectives. Empirically, SpidR-Adapt achieves rapid gains in phonemic discriminability (ABX) and downstream spoken language modeling scores (sWUGGY, sBLIMP, tSC), surpassing in-domain toplines after training on less than 1h of target-language audio and delivering $100\times$ greater data efficiency than standard multi-task training. These findings highlight a practical, architecture-agnostic path toward biologically inspired, data-efficient representations. We open-source the training code and model checkpoints at \url{https://github.com/facebookresearch/spidr-adapt}.
\end{abstract}

\section{Introduction}
Human infants demonstrate a remarkable capacity for language acquisition: at under 6-months of age, they begin specializing their perception of phonemic contrasts to the structures relevant to their native language \citep{eimas1971speech,Werker1984CrosslanguageSP,Kuhl2004EarlyLA}, all from continuous auditory input and with only 50 to 500 hours of speech exposure \citep{Bergelson2019,Cychosz2021,cristia2023systematic}.

In contrast, current self-supervised learning (SSL) models such as HuBERT \citep{hsu2021hubert} and WavLM \citep{chen2022wavlm} require thousands of hours of training data to learn meaningful linguistic representations, and even then, their learned units are brittle---sensitive to acoustic and contextual variability \citep{gat-etal-2023-augmentation,hallap2023evaluating}. When used as the basis for spoken language models (SLMs), these representations lead to limited language modeling performance compared to text-based systems \citep{lakhotia2021generative,hassid2023textually} and far worse than the learning trajectories of human infants \citep{bergelson2012}.

A key reason for this discrepancy lies in inductive biases: infants begin with strong predispositions for speech perception, such as sensitivity to phones, rhythmic regularities, and speaker-invariance \citep{kuhl1979speech,Kuhl2004EarlyLA,saffran2007infant}. These biases constrain learning to plausible linguistic structures, enabling rapid generalization from sparse input. By contrast, most machine learning systems are initialized from random weights and rely solely on statistical regularities of massive datasets. Without built-in inductive priors, they fail to discover linguistic abstractions of new languages efficiently.

To move toward the inductive efficiency of human learners, we propose a fast-adaptive self-supervised framework for speech representation learning including three broad components:
\begin{itemize}[itemsep=1pt, leftmargin=*]
\item \textbf{Multi-task Adaptive Pre-training (MAdaPT)}, a novel protocol that frames model learning as a bi-level optimization problem. The model is meta-optimized across several data-scarce adaptation episodes, each simulating a “lifetime” of low-resource language learning. Conceptually, this episodic design draws loose inspiration from evolutionary processes, with a second-order optimization occurring at an outer, population-like level that shapes the model’s inductive biases over generations. To further encourage cross-lingual abstraction, we introduce controlled active forgetting between episodes, resetting key model components to simulate the onset of a new  “lifetime,” thereby promoting robust, transferable representations.

\item \textbf{First Order Bi-level Optimization (FOBLO)}, a meta-optimization heuristic that efficiently approximates the second-order bi-level problem posed by MAdaPT. The inner loop trains the model to learn on unlabeled, under-resourced data, while the outer loop calibrates the meta-parameters using feedback from a gold-standard labeled set. 
    
\item \textbf{Interleaved light supervision}, which incorporates self-supervised training with occasional phone supervised steps, yielding an initialization that imitates human-robustness to contextual- and acoustic-variations of speech while being label-efficient.
\end{itemize}
Together, these mechanisms produce a model that achieves performance comparable to monolingual SSL systems trained on 6,000 hours of language data, despite seeing only 10 minutes to 100 hours of data in the target language. We further demonstrate that the resulting fast-adaptive model learns speech representations of an unseen language significantly faster than standard multi-task training.

We build on SpidR \cite{Poli2025SpidR}, a speech SSL model that achieves state-of-the-art (SOTA) performance on phonemic discrimination and SLM metrics with efficient training. Our framework extends SpidR with the above fast-adaptive components, yielding \OURS{}. Although our current implementation of MAdaPT-FOBLO uses SpidR as the backbone and focuses on speech representation, our framework is architecture-agnostic and broadly applicable to self-supervised models. Our results demonstrate a step toward data-efficient speech representation learning, conceptually motivated by the efficiency of early human language acquisition. 


\section{Related Works}

\subsection{Self-supervised learning}
Self-supervised learning has enabled speech models to learn rich representations from unlabeled audio, underpinning a wide range of downstream applications including ASR, emotion recognition, and spoken language modeling (SLM). Among these, SLM---where the objective is to capture linguistic structure directly from speech \citep{lakhotia2021generative,dunbar2021zero, borsos2022audiolm}---is particularly relevant for our work, given our motivation to build SSL models that enable human-like acquisition of spoken language. In the context of SLM, recent research has demonstrated that the semantic representativeness of learned units, in particular their phonemic discriminability, directly impacts downstream spoken language performance \citep{poli-etal-2024-improving,hallap2023evaluating}. Hence, in this work, when evaluating the performance of speech SSL models, we employ measures of phonemic discriminability such as ABX \cite{schatz2016abx}, PNMI \cite{hsu2021hubert}, and phone error rate.


Self-supervised models like HuBERT \citep{hsu2021hubert} and WavLM \citep{chen2022wavlm} use masked prediction and clustering to build speech representations, but require extensive training time. SpidR \cite{Poli2025SpidR} improves on prior SSL models by combining self-distillation and online clustering, achieving SOTA SLM results with more efficient training. This efficiency makes SpidR an ideal backbone for current meta-learning approaches.

\subsection{Meta-learning}
Meta-learning aims to optimize models for rapid adaptation to new tasks, often in low-resource settings \citep{finn2017model,nichol2018firstordermetalearningalgorithms}. This is typically achieved by performing two loops of optimization: in the inner loop, the model is repeatedly adapted to a new task, and in the outer loop, its meta-parameters are updated based on how well it adapts to that task. First-order model-agnostic meta-learning and Reptile \cite{nichol2018firstordermetalearningalgorithms}, in particular, use first-order outer loop updates, making them computationally attractive heuristics for large-scale meta-learning. 

Meta-learning has demonstrated significant effectiveness in improving out-of-domain (OoD) generalization. Recent studies have introduced risk-aware task selection frameworks that significantly improve adaptability and robustness without sacrificing training efficiency when facing distribution shifts \citep{wang2025model, qufast} while others have proposed meta-learning for OoD detection and model selection \citep{qin2024metaood}. In this paper, we evaluate generalization capability by meta-testing on OoD languages that are not available during meta-training.

Recent work has also explored active forgetting as a complementary mechanism for improving model plasticity \citep{chen2023languageplasticity, aggarwal2024exploring}. By periodically resetting parts of the model, such as embeddings or prediction layers, active forgetting encourages the formation of weights that can be reconfigured for new linguistic domains and prevents overfitting to unstable patterns. Here, we blend traditional meta-learning with active forgetting to amplify the adaptive benefits of both.

Despite their success in few-shot learning, meta-learning methods have seen limited application in speech models, where training typically relies on large, static corpora. Only a few studies explore meta-learning for speech classification or ASR \citep[e.g.,][]{chen2021userdefined,hsu2020metalowresource}, and none target self-supervised speech representations. In contrast, we apply meta-learning at the level of SSL itself for the goal of spoken language modeling. 

\begin{figure*}[ht!]
    \centering
    \includegraphics[width=0.85\textwidth]{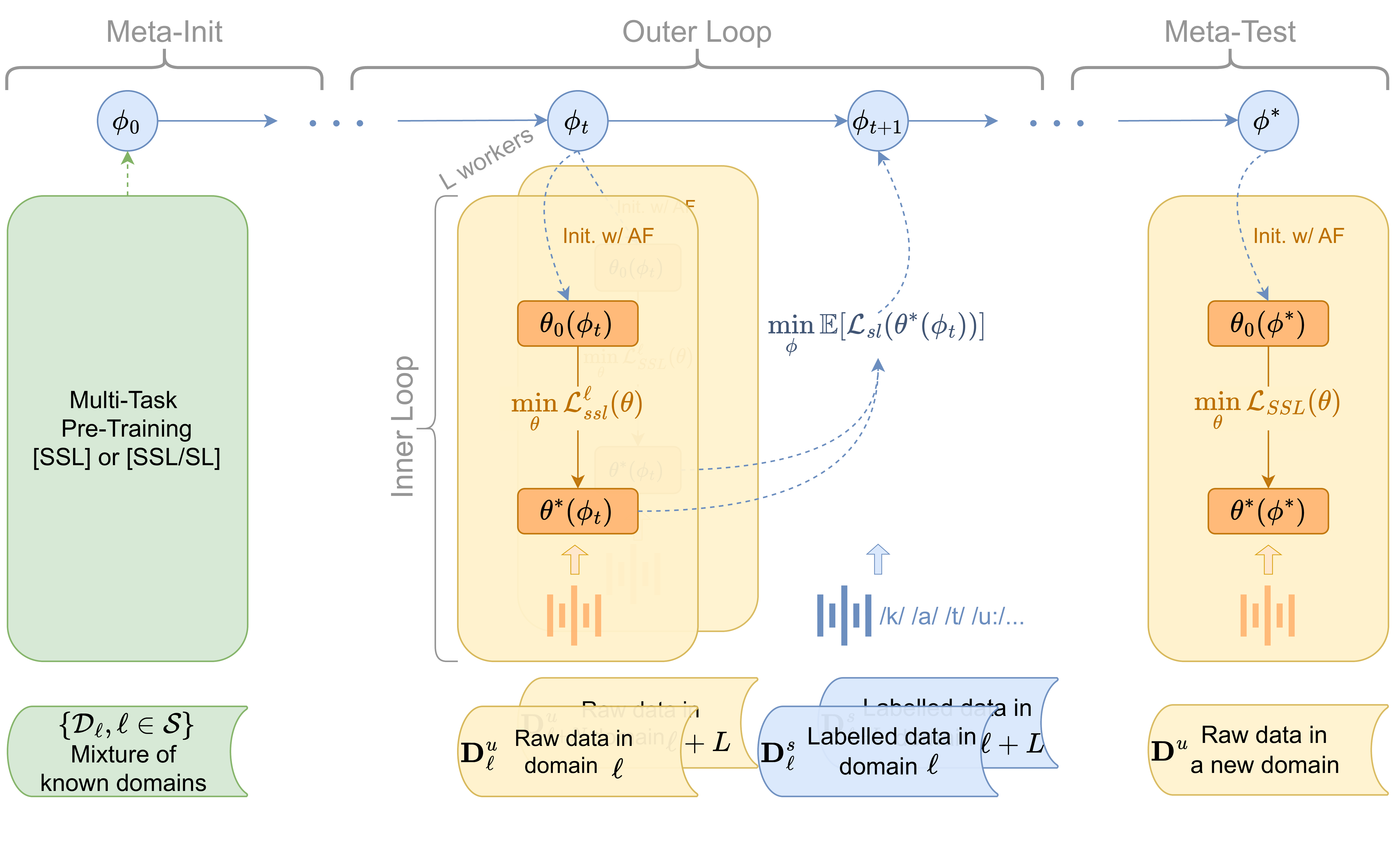}
    \caption{\textbf{Overview of \OURS{} for few-shot speech adaptation.} It consists of three main phrases: (1) meta-initialization performs multi-task pre-training with interleaved supervision, learning a robust initialization $\bm{\phi}_0$ from a mixture of source domains. (2) meta-training through MAdaPT-FOBLO optimizes this initialization for fast adaption to $\mathcal{D}_{\ell}$. Each worker conducts inner loop adaptation with active forgetting (AF) on unlabeled data, followed by outer loop updates that refine $\phi$ by minimizing the expected task loss on labeled data. (3) at meta-test time, the learned $\bm \phi^*$ is fast adapted to a new, unseen domain using only unlabelled data.}
    \label{fig:main}
\end{figure*}

\section{Methodology}
Here we introduce \OURS{}, a speech representation model tailored for rapid and robust adaptation to new languages with limited unlabeled audio data. First, we build a general multi-task training setup (\textbf{MAdaPT; Sec.~\ref{method:MAdaPT}}) that imitates fast-adaptation to new languages in low-resource scenarios, incorporating active forgetting to encourage stronger cross-lingual abstraction. This approach builds the adaptation process as a bi-level optimization problem. 
Then, to efficiently solve the nontrivial bi-level problem,  we introduce an empirical solution called first-order bi-level optimization (\textbf{FOBLO; Sec.~\ref{method:FOBLO}}), which avoids the heavy computational cost of second-order gradient steps in the outer loop. 
Finally, to stabilize meta-optimization, we propose initializing with a pre-trained model and design an interleaved supervised objective (\textbf{interleaved supervision; Sec.~\ref{method:light-weight-supervision-training}}).


\subsection{Multi-task Adaptive Pre-Training (MAdaPT)}
\label{method:MAdaPT}
The goal of MAdaPT is to address the OoD generalization challenge: the model is pre-trained on source (seen) linguistic domains with sufficient data and subsequently adapted on target (new) linguistic domains for which only limited unlabeled data is available. 

\paragraph{Notation.} Let $\mathcal{S}$ denote the set of source languages available during training and $\mathcal{T}$ represent the set of unseen target languages encountered during adaptation. For each source language $\ell \in \mathcal{S}$, we assume access to a sufficiently large unlabeled corpus $\mathcal{D}^{u}_\ell$ and, optionally, a small labeled corpus $\mathcal{D}^{s}_\ell$. In contrast, for each target language in $\mathcal{T}$,  only a limited unlabeled corpus is available. 


\paragraph{Episodic multi-lingual setup.} We cast the OoD challenge from seen to new languages as a meta-learning problem. To simulate fast adaptation to target languages with limited speech data, we partition the large unlabeled corpus $\mathcal{D}^{u}_\ell$ of each source language into multiple smaller data chunks $\{ \mathbf{D}^{u}_\ell \}$. Thus, one task in this work corresponds to a specific language $\ell$ and one scarce data chunk $\mathbf{D}^{u}_\ell$ as the training set. 
During meta-training, the model is presented with a mini-batch of task-specific episodes and is optimized in the outer loop based on learning performance of the inner loops. At the meta-test stage, we fine-tune the learned model on data-scarce tasks derived from each target language, evaluating adaptation in low-resource scenarios.

\paragraph{SpidR as backbone speech model.}
In this work, we deploy the SOTA speech representation model SpidR \cite{Poli2025SpidR} as our backbone, which has a teacher-student architecture. We represent it as
$\bm{\theta} = \big \{f(\cdot), E_{s}, E_{t}, \{{\bf{W}}^{k}\}, \{{\bf{C}}^{k} \} \big \}$, where $f(\cdot)$ is a convolutional downsampler, and $E_{s}$, $E_{t}$ are Transformer encoders for the student and teacher, respectively. The teacher is an exponential moving average of the student. ${\bf{W}}^k$ is the prediction head of the student and ${\bf{C}}^{k}$ is the target codebook of the teacher at the intermediate layer $k$ (where $L-K \leq  k \leq L$, with $L$ the number of Transformer layers and $K$ the number of codebooks).

Given a language $\ell$ with its low-resource dataset $\mathbf{D}^{u}_{\ell}$, we formalize the adaptation process as:
\begin{equation}
\label{eq:adapataion}
    \bm{\theta}^*_{\ell} = \arg \min_{\bm{\theta}} \mathcal{L}_{\text{ssl}}(\bm{\theta}; \mathbf{D}^{u}_{\ell}),
\end{equation}
where $\mathcal{L}_{\text{ssl}}$ denotes a self-supervised loss function, $\bm{\theta}$ represents all learnable parameters of the speech model SpidR, and $\bm{\theta}^*_{\ell}$ are the optimal model parameters specific to the language $\ell$. 
We note that $\mathbf{D}^{u}_{\ell}$ is not sufficient to train a specific speech model from scratch due to severe overfitting~\cite{dupoux2018cognitive}. 

\paragraph{Bi-level optimization.}
To mitigate model's overfitting to source languages, we propose a generic bi-level optimization framework which aims to learn meta-parameters from source languages that adapt rapidly to target languages. Within this framework, training with pure SSL in Equation~(\ref{eq:adapataion}) serves as an inner optimization; meanwhile, lightweight labeled data are deployed to supervise these adaptation processes at the outer level by shaping meta-parameters. Meta-parameters are shared across concurrent tasks and can be intuitively viewed as inductive biases for speech representation learning. Concretely, we interpret the learned meta-parameters $\phi$ as an adaptation prior: an initialization that can be efficiently reorganized toward the phonemic structure of a new language under limited unlabeled data.

For clarity, we instantiate meta-parameters $\bm{\phi}$ as the initial parameters of the backbone model in Equation~(\ref{eq:adapataion}).
Thus, the expected bi-level objective for \textbf{MAdaPT} is: 
\begin{equation}
\label{eq:framework}
\begin{aligned}
    & \min_{\bm{\phi}} \mathbb{E}_{\ell \sim \mathcal{S}} \big[ \mathcal{L}_{\text{sl}}(\bm{\theta}^*_\ell(\bm{\phi}); \mathbf{D}^{s}_\ell) ], \\
    & \quad \text{s.t.~} \bm{\theta}^*_\ell (\bm{\phi})  = \arg \min_{\bm{\theta}} \mathcal{L}_{\text{ssl}}(\bm{\theta}, \bm{\phi}; \mathbf{D}^{u}_\ell),
\end{aligned}
\end{equation}
where $\mathcal{L}_{\text{ssl}}$ denotes a self-supervised loss function in the inner level, performing adaptation from unlabeled speech data; and $\mathcal{L}_{\text{sl}}$ denotes a supervised loss function in the outer level. In contrast to regular meta-learning frameworks designed for supervised learning~\cite{finn2017model}, supervised information here is only used in the outer optimization while inner adaptations remain unsupervised. This 
preserves the assumption of low-resource, unlabeled data usage within the inner loop,  while leveraging supervised information in the outer loop to resolve the ambiguities of pure self-supervision.

\paragraph{Active forgetting in task adaptation.}
To suppress unstable and language-specific learning from past episodes, we introduce an active forgetting mechanism. During meta-training, SpidR's prediction heads and codebooks tend to be dominated by phonemic knowledge from source languages, hindering its generalization to new languages. 

To this end, we reinitialize these components at the start of each inner loop. Concretely, we copy the student and teacher parameters from the shared meta-parameters $\bm{\phi}$ as default but reset all heads and codebooks, yielding the optimization with initialization $\bm{\theta}_{\text{AF}}(\bm{\phi})$ for each inner loop at both meta-training and meta-test stages:
\begin{align}
\label{eq:active-forgetting}
    & \min_{\bm{\theta}} \mathcal{L}_{\text{ssl}}  ({\bm\theta}_{\text{AF}}( \bm{\phi}); \mathbf{D}^{u}_\ell), \text{ where} \\
    {\bm{\theta}}_{\text{AF}}( \bm{\phi}) & = \left\{ f(\bm{\phi}),\! E_{s}({\bm{\phi}}),\! E_{t}(\bm{\phi}),\! \{{\bf{W}}^{k}_0\},\! \{{\bf{C}}^{k}_0 \} \right\}. \nonumber
\end{align}

Each codebook ${\bf{C}}^{k}_0$ is sampled from a normal distribution $\mathcal{N}(0, 1)$ and each head ${\bf{W}}^{k}_0$ is warmed up for 20 steps using the first batch of $\mathbf{D}^{u}_\ell$. 


\subsection{First-Order Bi-Level Optimization (FOBLO)}
\label{method:FOBLO}

Solving the bi-level optimization in Equation~(\ref{eq:framework}) is non-trivial because both the inner and outer loops require multiple gradient steps. To make meta-training scalable, we introduce a first-order bi-level optimizer that yields a principled first-order approximation to the meta-gradient. In contrast to other first-order approximations~\cite{finn2017model, nichol2018firstordermetalearningalgorithms}, our optimizer is intended for a more challenging case where the inner and outer loops are served by different loss functions. 

Given a specific language $\ell$, the update of meta-parameters $\bm \phi$ can be formulated as: 
\begin{equation}
\label{eq:meta-update1}
   \bm \phi \leftarrow  \bm \phi - \beta \nabla_{\bm \phi} \mathcal{L}_{sl}(\bm{\theta}^*_\ell(\bm{\phi}); \mathbf{D}_{\ell}^{s}),
\end{equation} where $\beta$ is a learning rate in the outer loop used to update the meta-parameters $\bm \phi$.
Assume that the inner and outer loops perform $M$ and $N$ gradient steps, respectively. By applying chain rule to Equation (\ref{eq:meta-update1}) during backpropagation over $M$ inner steps, we can reformulate the meta-update as:
\begin{equation}
\label{eq:meta-update2}
\begin{split}
    \bm \phi \leftarrow \bm \phi - \beta \nabla_{\bm{\phi}} \mathcal{L}_{sl}(\bm{\theta}^{M}_\ell(\bm \phi); {\mathbf D}_{\ell}^{s}) 
    \\
    \cdot \prod_{m=1}^{M} \left[\mathbf{I} - \alpha \nabla_{\bm{\phi}^{m-1}_\ell} \left(\nabla_{\bm \phi}\mathcal{L}_{ssl}(\bm{\theta}^{m-1}_\ell(\bm \phi))\right)\right],
\end{split}
\end{equation} where $\alpha$ is the learning rate in the inner loop update and the task-specific parameters $\bm \theta_{\ell}^{m}$ denote the model's parameters after the $m^{th}$-inner step. 
To avoid the heavy computational cost in computing the Jacobian product of the second derivative in Equation~(\ref{eq:meta-update2}), we adopt a first-order approximation by dropping the second-order term (i.e., we stop the gradient through the inner loop). 

The outer loop typically performs $N$ supervised steps on labeled speech corpora ${\mathbf D}_{\ell}^{s}$. Following Reptile \cite{nichol2018firstordermetalearningalgorithms}, we approximate the outer loop gradient by the parameter difference between the end of the inner loop and the end of the outer loop:
\begin{equation}
\label{eq:meta-update3}
\nabla_{\bm{\phi}} \mathcal{L}_{sl}(\bm{\theta}^{M}_\ell(\bm \phi); {\mathbf D}_{\ell}^{s}) = \bm \theta_{\ell}^{M} - \bm \theta_{\ell}^{M+N}, 
\end{equation} where $\bm \theta_{\ell}^{M}$ is obtained after $M$ self-supervised inner steps starting from $\bm \theta$ and 
$\bm \theta_{\ell}^{M+N}$ is obtained by taking an additional $N$ supervised steps from $\bm \theta_{\ell}^{M}$. By substituting Equation~(\ref{eq:meta-update3}) into Equation~(\ref{eq:meta-update2}), \textbf{FOBLO} updates the meta-parameters as follows:
\begin{equation}
\label{eq:meta-update4}
   \bm \phi \leftarrow \bm \phi - \beta \mathbb{E}_{\ell \sim \mathcal{S}} [\bm \theta_{\ell}^{M} - \bm \theta_{\ell}^{M+N}].
\end{equation}
Both Reptile~\cite{nichol2018firstordermetalearningalgorithms} and the proposed FOBLO method use an outer loop learning rate $\beta$, but with different semantics: in Reptile, $\beta$ controls how far the meta-parameters move toward the task-adapted parameters along the full adaptation trajectory; in Equation~(\ref{eq:meta-update4}), $\beta$ controls how much the meta-parameters move in the direction of the supervised correction defined in Equation~(\ref{eq:meta-update3}), encouraging the model to meta-learn an initialization whose SSL adaptation aligns well with supervised targets.
Illustration of our work is provided in Figure~\ref{fig:main}. This work provides a principled and practical solution for few-shot self-supervised adaptation by nesting self-supervised inner loops within supervised outer loops.

\subsection{Interleaved Supervision}
\label{method:light-weight-supervision-training}
In practice, we find that initializing the meta-parameters from random weights leads to unstable learning dynamics and poor convergence (see Appendix~\ref{app:initialization}). Thus, to facilitate effective bi-level optimization, it is necessary to perform a dedicated pre-training phase prior to the meta-training stage.

To this end, we introduce an interleaved pre-training objective to obtain the most performative meta-initialization, denoted as $\bm \phi_0$. During the dedicated pre-training phase, we alternate between self-supervised and supervised objectives in an interleaved manner. This mechanism leverages both unlabeled and labeled data, allowing the model to benefit from large-scale unsupervised corpora while grounding representations with supervised signals. This pre-training objective is defined as:
\begin{equation}
\arg \min_{\bm \phi_0} \;
\left\{
\begin{aligned}
& \lambda \; \mathcal{L}_{\text{ssl}}(\bm \phi_0 ; \{\mathcal{D}^{u}_{\ell}\})  \\
& + (1-\lambda) \; \mathcal{L}_{\text{sl}}(\bm \phi_0; \{\mathcal{D}^{s}_\ell\})
\end{aligned}
\right.,
\label{eq:interleavedSL}
\end{equation}
where $\lambda \in \{0, 1\}$ is a binary hyperparameter.  Here, $\mathcal{L}_{\text{ssl}}$ denotes the self-supervised loss, applied to the union set of unlabeled corpora from all source languages, $\{\mathcal{D}^{u}_{\ell}, \ell \sim \mathcal{S}\}$; while $\mathcal{L}_{\text{sl}}$ is the supervised loss, applied to the union of labeled corpora $\{\mathcal{D}^{s}_{\ell}, \ell \sim \mathcal{S}\}$. 

 In the current work, we use two distinct meta-initializations: 1) \texttt{Multi-Task-PT~[SSL]}: setting $\lambda$ to $1$ throughout pre-training, corresponding to standard SSL; and 2) \texttt{Multi-Task-PT~[SSL/SL]} switching $\lambda$ to $0$ periodically, interleaving occasional supervised steps into the self-supervised training regime. The latter provides a stronger initialization for meta-training.

\section{Experiments}

We seek to address the following key questions:
(1) How data-efficient is \OURS{} in generalizing to the linguistic structure of new languages? 
(2) Can the MAdaPT framework lead to improvements when labeled data is unavailable during pre-training?
(3) Can \OURS{} produce improvements in downstream spoken language modeling? (4) Can \OURS{} outperform existing speech models under the OoD setup?

\begin{figure*}[ht!]
\centering \includegraphics{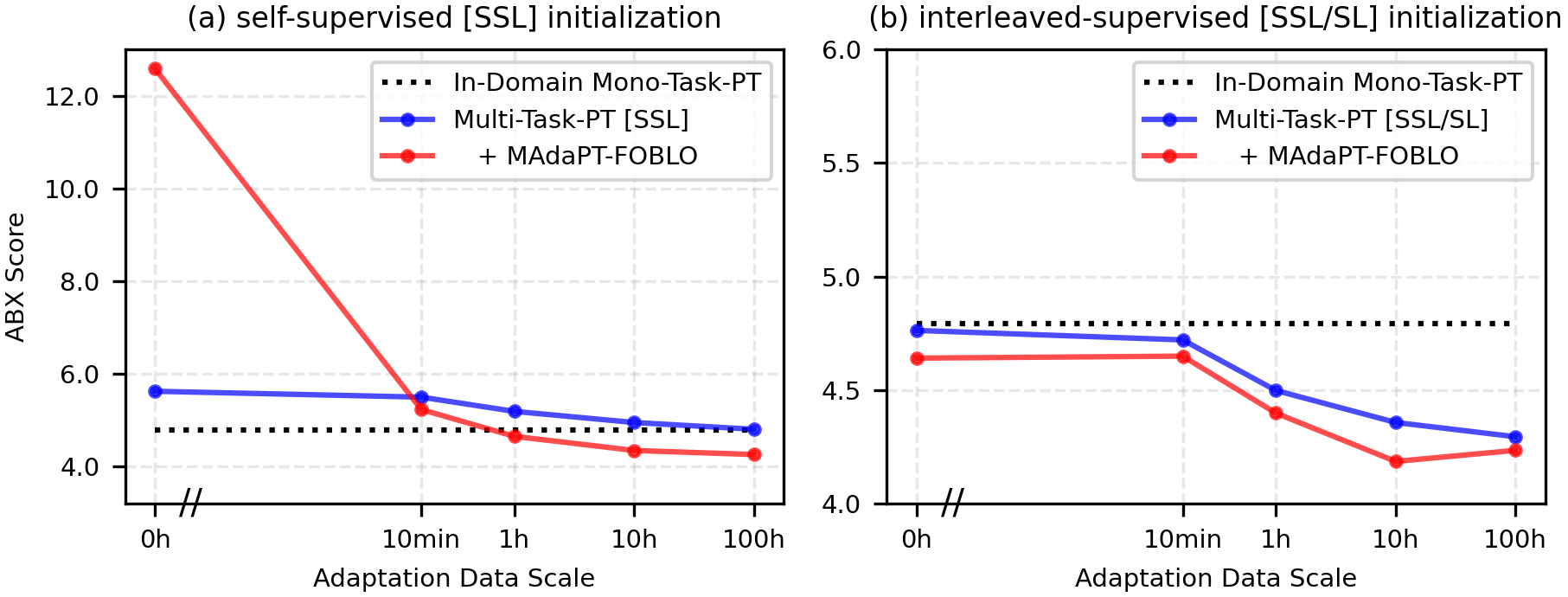}
\caption{\textbf{Data-efficiency of SpidR-Adapt on new languages across different adaptation data scales}. We report ABX scores (lower is better) averaged across three test languages (French, German, English) for two initialization strategies (a) self-supervision \texttt{[SSL]} and (b) interleaved-supervision \texttt{[SSL/SL]}. Each sub-figure compares our approach with the following: \texttt{\textcolor{black}{In-Domain Mono-Task-PT}}, the topline method, pretrained on 6k hours of in-domain data and  \texttt{\textcolor{blue}{Multi-Task-PT}}, standard multi-task pretraining baseline using \texttt{[SSL]} or \texttt{[SSL/SL]} regimes. By integrating the proposed solution, \texttt{\textcolor{red}{MAdaPT-FOBLO}}, with \texttt{Multi-Task-PT} as meta-initialization, we achieve highly efficient adaptation to new languages. For detailed results, refer to Appendix~\ref{app:detailed_abx}.
}
\label{fig:data-efficiency}
\end{figure*}


\paragraph{Datasets.} We collect data from 27 languages to evaluate adaptation capabilities of speech encoders under in-domain (ID) and out-of-domain (OoD) setups. 
We partition the languages as follows: 19 source languages for training; 5 target languages for development; and 3 target languages for testing. Importantly, there are no overlaps between source and target languages. 
Each source language is supported by a substantial unlabeled corpus (300 hours per language) collected from \texttt{VoxPopuli}~\cite{wang-etal-2021-voxpopuli} and a small phone-aligned corpus (maximum 50 hours per language) collected from \texttt{VoxCommunis Corpus}~\cite{ahn2022voxcommunis} to serve as labels for the FOBLO outer loop and for interleaved supervision.

Only small-scale unlabeled corpora are available for fast adaptation to target languages (mimicking infant learning settings). We construct four subsets per target language with durations 10 minutes, 1 hour, 10 hours, and 100 hours. To quantify the performance gap between ID and OoD training, we additionally collect large-scale in-domain training corpora from \texttt{VoxPopuli} for each test language. Each in-domain corpus comprises 6k hours---comparable in scale to the combined duration of the OoD corpora. The small-scale adaptation sets for these test languages are sampled from the same in-domain training pool; consequently, the OoD models are adapted using subsets of ID data. These choices are made to enable fair comparisons between ID and OoD models. 

Small-scale adaptation corpora were also created for the meta-development languages, sourced from \texttt{Common Voice}~\cite{Ardila2020CommonVoice} and used for model development. Further details on dataset construction are provided in Appendix \ref{app:dataset_details}.

\paragraph{Training Setup.} We perform multi-task pretraining of SpidR with self- or interleaved-supervised objectives (interleaving supervision every 10 steps; see Sec.~\ref{method:light-weight-supervision-training}). These models serve as initializations for meta-training wherein we train across 800 episodes, each episode consisting 1800 inner and 200 outer steps. In each inner loop, the model is trained on a random 10 hour data chunk of a random source language. Training is performed across 16 GPUs in a distributed fashion. Details regarding training can be found in Appendix \ref{app:training_details}.

\subsection{Data-Efficiency When Adapting on New Languages}
\label{experiment:data_efficiency}
To evaluate data efficiency, we adapt meta-trained models to new target languages using only limited unlabeled data. We benchmark our approach against baselines using ABX (lower is better), computed with the fastabx toolkit~\cite{poli2025fastabx}.

ABX scores quantify how well model embeddings capture phone distinctions and correlate strongly with downstream SLM performance \cite{Poli2025SpidR}, serving as an efficient zero-shot proxy. In the ABX task, embeddings are computed for three triphones: \textit{A}, \textit{B}, and \textit{X}. Here, \textit{A} and \textit{X} are instances of the same triphone, while \textit{B} differs in its central phone (e.g., \textipa{/bag/} vs. \textipa{/beg/}). The model succeeds if \textit{X} is closer to \textit{A} than to \textit{B} in embedding space. The within-speaker condition uses triphones from the same speaker, while the across-speaker condition uses \textit{A} and \textit{B} from one speaker and \textit{X} from another, making the task more challenging. 

Figure~\ref{fig:data-efficiency} shows results: the x-axis indicates adaptation data size, and the y-axis shows ABX scores averaged across our three test languages and between within- and across-speaker ABX conditions (individual trends are consistent). Using SpidR as the backbone under two meta-initialization strategies, self- and interleaved-supervised initialization (Sec.~\ref{method:light-weight-supervision-training}), we compare: 
1) \textbf{In-Domain Mono-Task-PT}: Standard in-domain pre-training on sufficient data (6k hours) from the target language, serving as topline. Since every small-scale evaluation subset is drawn from the in-domain training pool, we do not perform additional small-scale adaptation: In-Domain PT therefore appears as a horizontal line in Figure~\ref{fig:data-efficiency}.
2) \textbf{Multi-Task-PT}: Standard OoD pre-training with ample unlabeled data from all source languages, using the same data-feeding protocol as In-Domain PT. This serves as our primary OoD baseline.
3) \textbf{MAdaPT-FOBLO}: Our proposed approach, combining MAdaPT with its first-order approximation FOBLO. When used with SpidR as the backbone and with interleaved-supervised initialization, this constitutes our few-shot learning speech encoder, \OURS{}. 

Figure~\ref{fig:data-efficiency}~(a) shows that Multi-Task-PT underperforms In-Domain PT especially when the adaptation budget is small ($<100~\text{hours}$). This suggests that regular multi-task pre-training lacks the adaptation capacity needed for unseen targets, and simply mixing several source languages during pre-training does not guarantee better generalization. 

In contrast, MAdaPT-FOBLO improves rapidly, demonstrating strong adaptability to OoD data. Notably, with just 1 hour of unlabeled target-language audio, MAdaPT-FOBLO reaches parity with In-Domain PT---a $100\times$ improvement in data efficiency over Multi-Task-PT. This efficiency is critical for real-world scenarios where language corpora are scarce.

As shown in Figure~\ref{fig:data-efficiency}~(b), the interleaved-supervised initialization (Multi-Task-PT [SSL/SL]) provides a better starting point (lower initial ABX) than self-supervised initialization (Multi-Task-PT [SSL]). However, regardless of initialization, the incorporation of MAdaPT-FOBLO delivers the largest gains in rapid adaptation to unseen languages. This suggests that while initialization can set a stronger baseline, the adaptation strategy is the primary driver of sustained performance improvements.

We also note an important stability–plasticity tradeoff: MAdaPT-FOBLO has a weaker zero-shot performance, reflected in the high $0$ hour ABX scores under SSL initialization. This is a caveat of the current formulation, since Equation~(\ref{eq:framework}) optimizes $\phi$ exclusively for post-adaptation performance without explicitly preserving zero-shot behavior. However, this also reveals that strong plasticity does not necessarily entail strong zero-shot performance. MAdaPT-FOBLO (with SSL initialization) yields a model that reorganizes more effectively once even minimal unlabeled data is available ($\geq 10$ minutes), consistently surpassing Multi-Task-PT baselines.

\begin{table}
    \centering
    \caption{\textbf{Comparisons with MAdaPT-Reptile}, a purely SSL solution for MAdaPT. 
    ABX scores averaged across $10$ minutes to $100$ hours training (excluding zero-shot, $0$ hours). Although Reptile under-performs FOBLO, it achieves better results than baseline Multi-Task-PT, demonstrating the effectiveness of MAdaPT. 
    \label{tab:reptile}}
    \begin{tabular}{l|cc}
\toprule
\multirow{2}{*}{\textbf{Method}} 
    & \multicolumn{2}{c}{\textbf{Avg. ABX (w/o 0h)} $\downarrow$} \\
    &  \makecell{\textit{Within-}\\\textit{Speaker}} & \makecell{\textit{Across-}\\\textit{Speaker}} \\
\midrule
Multi-Task-PT~[SSL] & 4.33  & 5.89\\
~~~ + MAdaPT-Reptile & \underline{4.19} & \underline{5.59} \\
\rowcolor{gray!10}
~~~ + MAdaPT-FOBLO & \textbf{4.01} & \textbf{5.24}\\
\bottomrule
\end{tabular}
\end{table}

\begin{table*}
    \centering
    \caption{\textbf{Spoken language modeling results (in \%) of the English-adapted models.} We report the average SLM metrics, including sWUGGY, sBLIMP, and tSC (higher is better). Across both meta-initalizations, MAdaPT-FOBLO consistently outperforms the Multi-Task-PT baseline, surpassing the In-Domain topline. MAdaPT-Reptile comes a close second. The best results are shown in \textbf{bold}, and second-best are \underline{underlined}.  For detailed results, refer to Appendix~\ref{app:slm-details}.
    \label{tab:slm}}
    \renewcommand{\arraystretch}{1.2}
\setlength{\tabcolsep}{10pt}
\begin{tabular}{lcccccc}
\toprule
\textbf{Method} & \textbf{0h} & \textbf{10m} & \textbf{1h} & \textbf{10h} & \textbf{100h} & \textbf{Avg. (w/o 0h) $\uparrow$} \\
\midrule
In-Domain Mono-Task-PT                  &  \multicolumn{6}{c}{\textit{$6000$ hours in-domain training; Topline score is $65.27$.}}\\
\midrule
Multi-Task-PT~[SSL]  & 63.49 & 63.80 & 64.63 & 64.51 & 65.20 & 64.54 \\
~~~~~~~ + MAdaPT-Reptile          & 64.42 & 64.42 & 64.47 & 64.59 & 64.72 & 64.55 \\
\rowcolor{gray!10}
~~~~~~~ + MAdaPT-FOBLO            & 59.08 & 65.30 & 65.19 & 65.73 & 66.39 & 65.65 \\
Multi-Task-PT~[SSL/SL]            & 64.72 & 65.34 & \underline{65.77} & 66.02 & 66.07 & 65.80 \\
~~~~~~~ + MAdaPT-Reptile          & \textbf{65.73} & \underline{65.44} & 66.02 & \textbf{66.77} & \textbf{67.00} & \underline{66.31} \\
\rowcolor{gray!10}
~~~~~~~ + MAdaPT-FOBLO            & \underline{65.31} & \textbf{65.79} & \textbf{66.32} & \underline{66.55} & \underline{66.85}  & \textbf{66.38} \\
\bottomrule
\end{tabular}
\end{table*}

\subsection{MAdaPT for Pure Self-Supervision}
\label{experiment:reptile}

Here we consider an extreme setting in which no supervised training data is available for source languages. In this regime, MAdaPT must be optimized using a purely self-supervised procedure. 

To instantiate MAdaPT without labels, we adopt Reptile~\cite{nichol2018firstordermetalearningalgorithms}, a first-order meta-learning heuristic that approximates the meta-gradient assuming identical inner and outer loop objectives. Here, the meta-update is written as $\bm\phi \leftarrow  (1- \beta) \bm\phi - \beta \mathbb{E}_{\ell \sim \mathcal{S}} [ \bm\theta^{M+N}_{\ell}]$, where \(\beta\) trades off the previous meta-parameters and the task-specific solution after each episode. In contrast to our FOBLO update in Equation (\ref{eq:meta-update4}), $\bm \theta^{M+N}_{\ell}$ in Reptile denotes parameters obtained by pure self-supervised training for a total of \(M{+}N\) steps.

In Table~\ref{tab:reptile}, we report ABX (in \%) averaged over adaptation budgets from \(10\) minutes to \(100\) hours and three test languages. The results emphasize that MAdaPT-based optimization consistently improves over standard Multi-Task-PT, with FOBLO (which requires supervised labels) achieving the strongest performance. Notably, when all source languages lack supervision, Reptile, used as a purely self-supervised instantiation of MAdaPT, outperforms the baseline Multi-Task-PT.
These findings underscore the importance of a tailored multi-task framework for low-resource OoD adaptation.

\subsection{Evaluating Downstream Spoken Language Models}
\label{experiment:slm}

We use OPT-1.3B \cite{zhang2022optopenpretrainedtransformer} as SLM backbones to evaluate the language modeling performance of SSL models adapted on English test sets, using three complementary linguistic metrics.
1) \textbf{Lexical: sWUGGY} ~\citep{nguyen2020zeroresourcespeechbenchmark} tests whether the model assigns higher probability to true words than to matched non-words.
2) \textbf{Syntax: sBLIMP} requires the model to choose grammatical sentences from minimal pairs.
3) \textbf{Discourse/Narrative: Spoken Topic StoryCloze}~\citep{mostafazadeh-etal-2017-lsdsem} asks the model to select appropriate continuations for short stories. We report accuracy (in \%) averaged across the three metrics in Table~\ref{tab:slm}. Detailed per-task results are included in the Appendix~\ref{app:slm-details}.

Table~\ref{tab:slm} shows that MAdaPT-FOBLO achieves rapid gains under the few-shot adaptation scenario (for both SSL and SSL/SL initializations). MAdaPT-Reptile comes a close second, with especially strong zero-shot performance.

\begin{table*}[ht]
    \centering
    \caption{\textbf{Results on phoneme discovery.} We present the scores on the DiscoPhon benchmark after finetuning models on 10 hours of language data and mapping their 256 speech units to the ground truth phonemes. Models were individually finetuned on one of six test languages and aggregate results across languages is presented here. MAdaPT-FOBLO outperforms alternative speech SSL models (HuBERT, DinoSR) and is on par with MAdaPT-Reptile. Both MAdaPT-FOBLO and MAdaPT-Reptile are initialized using Multi-Task-PT [SSL/SL] here. Complete results are provided in Appendix~\ref{app:detailed_pd_benchmark}. Best scores (in \%) are in \textbf{bold} and second best are \underline{underlined}.}
    \label{tab:comparisons}
    \renewcommand{\arraystretch}{1.1}
\setlength{\tabcolsep}{10pt}
\begin{tabular}{lc|ccccc}
\toprule
 \textbf{Method} & \textbf{\makecell{Backbone \\\ Model}} & \textbf{PER $\downarrow$} & $\bm{R}$\textbf{-value} $\uparrow$ & $\bm{F_1} \uparrow$ & \textbf{PNMI $\uparrow$} & \textbf{ABX $\downarrow$} \\
\midrule
Multi-Task-PT [SSL] & HuBERT & 98.02 & 24.37 & 61.16 & 57.56 & 5.49 \\
Multi-Task-PT [SSL] & DinoSR & 69.79 & 43.96 & 66.63 & 65.91 & 6.64\\
Multi-Task-PT [SSL] & SpidR & 48.63 & 61.84 & 71.10 & 69.34 & 4.47 \\
\midrule
MAdaPT-Reptile & SpidR  & \underline{37.05} & 
\textbf{75.01} &  \textbf{78.14} & \underline{69.42} & \underline{4.11} \\
\rowcolor{gray!10}
MAdaPT-FOBLO & SpidR  & \textbf{36.58} & \underline{73.91} & \underline{77.51} & \textbf{71.64} & 
\textbf{4.10} \\
\bottomrule
\end{tabular}
\end{table*}

\subsection{Evaluating on Phoneme Discovery}
To further investigate the adaptability of the proposed methods, we compare them with performant speech SSL models, HuBERT \cite{hsu2021hubert} and DinoSR \cite{liu2023dinosr} trained under the OoD Multi-Task-PT setup, on the DiscoPhon benchmark \cite{Poli2026Phoneme}. This benchmark targets the automatic discovery of phoneme inventories from limited raw speech (10 hours maximum, in 6 development and 6 test languages). Models are evaluated by mapping the discrete speech units to their most frequently associated phonemes prior to evaluation (fixing the number of units either to 256 in the many-to-one setting or to the ground truth number of phonemes in the one-to-one setting). Evaluating our models in the many-to-one setting, we report the following metrics: 1) phone error rate (\textbf{PER}); 2) $\bm{R}$\textbf{-value} \cite{rasanen09b_interspeech} and $\bm{F_1}$ segmentation scores; 3) phone-normalized mutual information (\textbf{PNMI}), measuring the uncertainty about a phone label eliminated by a predicted unit; 4) \textbf{ABX} on continuous representations (within- and across-conditions). As shown in Table~\ref{tab:comparisons}, \OURS{} (i.e., MAdaPT-FOBLO) consistently outperforms the alternative speech SSL models (HuBERT, DinoSR) with the alternative meta-learning framework (Reptile) also demonstrating strong performance. Full results are reported in Appendix~\ref{app:detailed_pd_benchmark}.

\section{Conclusion}
We present \OURS{}, a speech representation model that enables data-efficient adaptation to new languages by combining meta-adaptive pretraining, bi-level optimization, and interleaved supervision. Improving over in-domain model performance with as little as 1 hour of target-language audio, \OURS{} is over $100\times$ more data efficient than traditional multi-task methods and demonstrates the effectiveness of a tailored meta-learning framework for flexible representation learning in low-resource settings.

\section*{Limitations}
This work offers promising data-efficiency in few-shot speech representation learning, but several limitations remain. Model performance is influenced by the choice of meta-initialization, suggesting that further research is needed into more robust meta-learning that can be trained without meta-initialization. Supervised information from source languages is still required at the outer-level, which limits scaling of source languages. Further hyperparameter exploration is needed to identify superior configurations of SpidR-Adapt---some initial explorations in this direction are summarized in Appendix~\ref{app:extended}. Additionally, training of spoken language models has not been included into the meta-learning framework and hence is not data-efficient; future work could focus on applying meta-learning directly to SLM training to enhance efficiency and reduce data requirements.

\section*{Acknowledgements} 
 MP acknowledges PhD funding from Agence de l’Innovation de Défense and HPC resources from GENCI-IDRIS (Grant 2023-AD011014368). ED in his EHESS role was supported in part by the Agence Nationale pour la Recherche (ANR-17-EURE0017 Frontcog, ANR10-IDEX-0001-02 PSL*) and an ERC grant (InfantSimulator). Views and opinions expressed are those of the authors only and do not necessarily reflect those of the European Union or the European Research Council. Neither the European Union nor the granting authority can be held responsible for them. YLC contributed to this work while at Meta.

\newpage
\bibliography{custom}
\appendix
\newpage
\section*{Appendix}

\label{appendix}

\section{Details of Datasets}
\label{app:dataset_details}

\begin{table*}[ht]
    \centering
    \caption{\textbf{Summary of unlabeled datasets utilized across training and evaluation.} Data was accumulated from \texttt{VoxPopuli} (VP; \citeauthor{wang-etal-2021-voxpopuli}, \citeyear{wang-etal-2021-voxpopuli}) corpus and \texttt{Common Voice} (CV; \citeauthor{Ardila2020CommonVoice}, \citeyear{Ardila2020CommonVoice}).}
    \label{tab:datasets}
    \begin{tabular}{c|c|cc}
\toprule
\multirow{2}{*}{\textbf{Split}} & \multirow{2}{*}{\textbf{In-Domain Training}} & \multicolumn{2}{c}{\textbf{Out-of-Domain Training}} \\
    & & \textbf{Pre-Training} & \textbf{Fast Adaptation} \\
\midrule
\multirow{6}{*}{Dev.} & \multirow{6}{*}{\shortstack{In-domain Training \\ not performed}} 
    & \textit{\underline{~5700 hours}} & \textit{\underline{10 minutes, 1 hour, 10 hours}} \\
    & & \multirow{5}{*}{\makecell{VP 19 langs. \\\ (w/o target \\\ adaptation langs.)}} & CV Swahili \\
    &  &  & CV Tamil \\
    &  &  & CV Thai \\
    &  &  & CV Turkish \\
    &  &  & CV Ukrainian \\
\midrule
\multirow{4}{*}{Test}
    & \textit{\underline{6000 hours}} & \textit{\underline{~5700 hours}} & \makecell{\textit{\underline{10 minutes, 1 hour, 10 hours, 100 hours}} \\ \textit{(subset of In-Domain Training set)}} \\
    & VP English & \multirow{3}{*}{\makecell{VP 19 langs. \\\ (w/o target \\\ adaptation langs.)}} & VP English \\
    & VP French &  & VP French \\
    & VP German &  & VP German \\
\bottomrule
\end{tabular}
\end{table*}

Table~\ref{tab:datasets} summarizes the unlabeled datasets used for meta-training, meta-development, and meta-testing. To prepare the unlabeled training data, we apply the \texttt{Silero Voice Activity Detector} to the audio files, segmenting them into smaller audio files ranging from $0.5$ to $30$ seconds in duration (with mean $14.6$ seconds). This pre-processing step ensures that the model is exposed to realistic, variable-length speech segments during both training and evaluation.
For reproducibility, the start and end timestamp metadata for all processed audio files used in training and evaluation are available in the accompanying GitHub repository.

In addition to the unlabeled dataset, we also use a small supervised dataset, mainly sourced from \texttt{VoxCommunis Corpus}~\cite{ahn2022voxcommunis}. This corpus comprises phoneme alignments inferred on \texttt{Common Voice} \cite{Ardila2020CommonVoice} data using Montreal Forced Aligners (MFA; \citeauthor{mcauliffe2017montreal}, \citeyear{mcauliffe2017montreal}). While \texttt{CommonVoice} has data for 18 training languages, it does not contain data for one language---Croatian. To obtain a labeled set in Croatian, we use the transcribed set of \texttt{VoxPopuli} \cite{wang-etal-2021-voxpopuli} and align phonemes using off-the-shelf MFA models.  We clean the alignment data by applying similar filtering and phoneme mapping measures employed by \citet{Ortiz2025MAUBERT}. This includes filtering out alignments with \texttt{spn} segments or with non-silent phones that are excessively long (which indicate alignment errors), fixing diacritics that were wrongly attached to adjacent phones, and replacing some MFA phones with their IPA equivalents (\textipa{[\textg]} becomes \textipa{[g]}). The amount of phoneme-aligned data available varied widely based on language---to avoid overfitting on any one language, we limit the maximum quantity to 50 hours per language, which results in a labeled dataset of 372 hours.

To compute ABX scores on test languages in experiments Sec.~\ref{experiment:data_efficiency} and Sec.~\ref{experiment:reptile}, we use phoneme alignments obtained from the test set of the \texttt{Zero Resource 2017 Challenge} \cite{dunbar2017zero}. For the development languages, we use data from \texttt{Common Voice} \cite{Ardila2020CommonVoice} and alignments from \texttt{VoxCommunis} \cite{ahn2022voxcommunis}. 


\section{Details of Training Setup}
\label{app:training_details}
\subsection{Pre-training}
Models are trained using a distributed setup across 16 GPUs. Default SpidR hyperparameters \cite{Poli2025SpidR} are used for pre-training the ID mono-task and the OoD multi-task models. In interleaved supervised pre-training (i.e., Multi-Task-PT [SSL/SL]), every tenth step is backpropagated using phone supervised loss (hence in equation \ref{eq:interleavedSL}, $\lambda = 0 \text{ if } \text{step} \bmod 10 = 0, \text{ else } 1$). For prediction of supervised labels, language-specific classifier heads (19 heads in total) are attached to the $8^{th}$ transformer layer of the SpidR model. Here, the $8^{th}$ layer was used because exploration of hyperparameters indicated it as being optimal for few-shot performance on developmental languages. During supervised training steps, utterances are batched by language; while during self-supervised training steps, each batch consists of a mix of languages. In self-supervised pre-training (i.e., Multi-Task-PT [SSL]), standard SSL loss (as defined by the SpidR architecture) is used throughout. The OoD multi-task models trained under these schema are used as initialization weights for meta-training.

\subsection{Meta-Training}

Eight MAdaPT episodes are trained in parallel across 16 GPUs. During meta-training, each episode consists of 2,000 steps, with 1,800 steps for the inner loop (self-supervised adaptation) and 200 steps for the outer loop (supervised meta-optimization). For each inner loop task $\mathbf{D}^{u}_\ell$, we use a randomly chosen 10-hour data chunk from a randomly chosen source language. For the outer loop otimization, the inner loop language $\ell$ is retained, but data duration is not fixed at 10 hours. The overall training spans 200,000 steps, resulting in a total of 800 episodes (calculated as $200,000/2,000\times8=800$ episodes). This meta-training setup is chosen for both practicality of implementation (on limited compute and with limited time) and to closely mimic the low-resource adaptive fine-tuning scenario central to our research.

For the self-supervised initialization of FOBLO, the supervised outer loop optimization is applied to the $6^{th}$ layer of the model; while, for the interleaved-supervised initialization, it is applied to the $8^{th}$ layer (staying consistent with the supervised layer during meta-initialization). The FOBLO supervised layers are selected based on best performing layers of the meta-initialization models computed on the development language set. When computing ABX scores, we thereby report results from the $6^{th}$ and $8^{th}$ layers for the self- and interleaved-supervised models, respectively. For Reptile, since no layer-specific optimization is employed, we identify the best-performing layers of models through exploration on the development set, finding $6^{th}$ and $8^{th}$ layers to be optimal for the self- and interleaved-supervised models, respectively, across all adaptation scales.

In SpidR, the teacher is trained as an exponential moving average of the student, with the decay of the teacher at the timestep $t$ defined as $1 - (1 - \beta_0) \exp(- t / T)$. We find that some meta-training configurations (specifically, trainings initialized using interleaved supervision or meta-trained using FOBLO) perform better when trained with $\beta_0=1$, effectively producing a frozen teacher. Hence, we select the best performing value of $\beta_0$ (from $1.0$ and the default $0.999$) for each meta-training variant (i.e., Reptile or FOBLO with SSL or SSL/SL initializations) based on few-shot performance on the development language set.

\begin{figure*}[ht]
\centering \includegraphics[width=\linewidth]{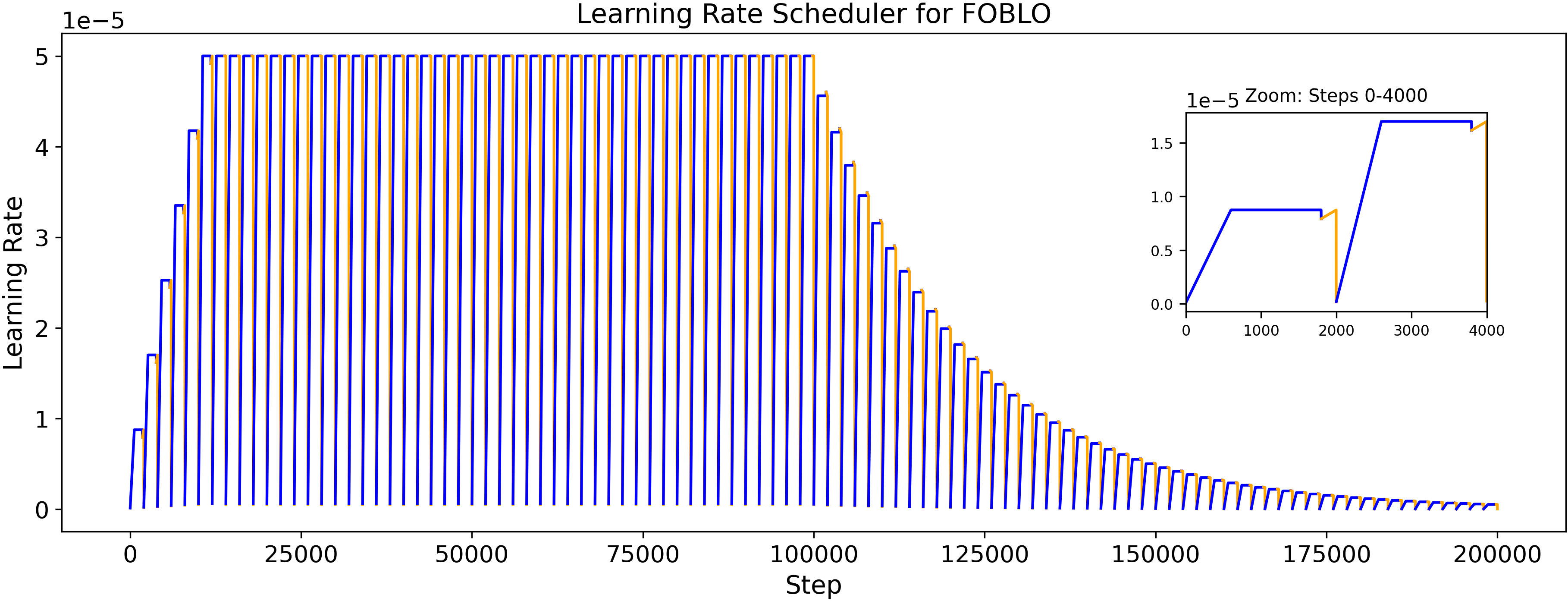}
\caption{\textbf{Learning rate scheduler for FOBLO.} We use \textcolor{blue}{blue} and \textcolor{orange}{orange} to represent the learning rate for self-supervised inner steps and supervised outer steps, respectively. The overall training has 200,000 steps. The learning rate scheduler alternates between inner loop and outer loop steps within each episode, with resets every 2,000 steps. The inner loop uses a constant rate after a warmup, while the outer loop follows a tri-stage schedule.}
\label{fig:learning_rate_curve}
\end{figure*}

Within each meta-training inner loop, we use a constant learning rate adding a small warmup for 600 timesteps at the beginning of each loop. The learning rate within each episode is identified through a tri-stage learning rate scheduler with maximum learning rate of \num{5e-5}. The detailed scheduler is illustrated in Figure~\ref{fig:learning_rate_curve}.

\subsection{Fast Adaptation Training}
For fast adaptive fine-tuning to the OoD target languages, we use a single GPU. For each model variant (i.e., Multi-Task-PT, MAdaPT-Reptile, or MAdaPT-FOBLO with SSL or SSL/SL initializations)  and each adaptation dataset size (10 minutes  to 100 hours), we conduct a hyperparameter exploration on the development language set to identify optimal training timesteps (varied between 4,000 and 24,000), learning rate (constant learning rate of \num{5e-4} or \num{5e-5}), and $\beta_0$ for the teacher decay ($1.0$ or default $0.999$). The best checkpoint for each adaptation run is selected based on the lowest validation loss, ensuring optimal model performance for downstream evaluations.

\section{Detailed Experimental Results of the Main Manuscript}

\subsection{Detailed Results of ABX scores}
\label{app:detailed_abx}
Here, we present detailed ABX scores for both within-speaker and across-speaker setups as illustrated in Figure~\ref{fig:data-efficiency}. As shown in Table~\ref{tab:details-data-3test}, the In-Domain Mono-Task-PT [SSL] models are trained with sufficient in-domain data (6k hours per language), and hence are the toplines. Moreover, we evaluate all methods on the five development languages, with their ABX scores reported in Table~\ref{tab:details-data-5dev}. Due to the lack of unlabeled corpora for these five development languages, the in-domain topline performance is not reported in the table. Across both tables, our proposed MAdaPT-FOBLO consistently outperforms the multi-task baseline and achieves performance comparable to the MAdaPT-Reptile method. Notably, when self-supervised initialization is applied, our approach rapidly improves performance as adaptation time increases, highlighting its data efficiency and overall effectiveness.

\begin{table*}[ht]
    \centering
    \caption{\textbf{Detailed \textit{Within-Speaker} and \textit{Across-Speaker} ABX scores (in \%) on \textcolor{blue}{$3$ TEST} languages}. MAdaPT-FOBLO and MAdaPT-Reptile with SSL/SL regimes show superior performance, surpassing In-Domain Mono-Task-PT with limited data. The best scores are in \textbf{bold} and second best are \underline{underlined}.} 
    \label{tab:details-data-3test}
    \renewcommand{\arraystretch}{1.1}
\begin{tabular}{lcccccc}
\toprule
\textbf{Method} & \textbf{0h} & \textbf{10m} & \textbf{1h} & \textbf{10h} & \textbf{100h} & \makecell{\textbf{Avg.} \\\ \textbf{(w/o 0h) $\downarrow$}} \\
\midrule
\multicolumn{7}{c}{\underline{\textit{Within-Speaker ABX} }} \\
In-Domain Mono-Task-PT  &  \multicolumn{6}{c}{\textit{$6000$ hours training; Topline score is $4.10$.}}\\
\midrule
Multi-Task-PT~[SSL]       & 4.65 & 4.56 & 4.40 & 4.23 & 4.13 & 4.33 \\
~~~~~~~ + MAdaPT-Reptile  & 4.50 & 4.34 & 4.29 & 4.10 & 4.03 & 4.19 \\
\rowcolor{gray!10}
~~~~~~~ + MAdaPT-FOBLO    & 10.05 & 4.51 & 4.05 & \underline{3.78} & \underline{3.69} & 4.01 \\
Multi-Task-PT~[SSL/SL]    & 4.10 & 4.08 & 3.94 & \underline{3.78} & 3.71 & 3.88 \\
~~~~~~~ + MAdaPT-Reptile  & \textbf{3.89} & \textbf{3.94} & \textbf{3.82} & \textbf{3.62} & \textbf{3.66} & \textbf{3.76} \\
\rowcolor{gray!10}
~~~~~~~ + MAdaPT-FOBLO    & \underline{4.00} & \underline{4.07} & \underline{3.84} & \textbf{3.62} & 3.70 & \underline{3.80} \\
\midrule
\multicolumn{7}{c}{\underline{\textit{Across-Speaker ABX}}} \\
In-Domain Mono-Task-PT  &  \multicolumn{6}{c}{\textit{$6000$ hours training; Topline score is $5.47$.}}\\
\midrule
Multi-Task-PT~[SSL] & 6.60 & 6.44 & 5.99 & 5.68 & 5.48 & 5.89 \\
~~~~~~~ + MAdaPT-Reptile & 5.97 & 5.82 & 5.72 & 5.45 & 5.38 & 5.59 \\
\rowcolor{gray!10}
~~~~~~~ + MAdaPT-FOBLO & 15.12 & 5.96 & 5.26 & 4.92 & 4.83 & 5.24 \\
Multi-Task-PT~[SSL/SL] & 5.42 & 5.36 & 5.06 & 4.93 & 4.88 & \underline{5.06} \\
~~~~~~~ + MAdaPT-Reptile & \textbf{5.19} & \textbf{5.16} & \underline{4.97} & \underline{4.78} & \underline{4.79} & \textbf{4.93} \\
\rowcolor{gray!10}
~~~~~~~ + MAdaPT-FOBLO & \underline{5.28} & \underline{5.23} & \textbf{4.96} & \textbf{4.76} & \textbf{4.77} & \textbf{4.93} \\
\bottomrule
\end{tabular}
\end{table*}

\begin{table*}[ht]
    \centering
    \caption{\textbf{Detailed \textit{Within-Speaker} and \textit{Across-Speaker} ABX scores (in \%) on \textcolor{blue}{$5$ DEVELOPMENT} languages}. MAdaPT-FOBLO outperforms alternate methods in phoneme representation. Hyperparameters are tuned using results from the development language set. The best scores are in \textbf{bold} and second best are \underline{underlined}.}
    \label{tab:details-data-5dev}
    \renewcommand{\arraystretch}{1.2}
\begin{tabular}{lcccccc}
\toprule
\textbf{Method} & \textbf{0h} & \textbf{10m} & \textbf{1h} & \textbf{10h} & \makecell{\textbf{Avg.} \\\ \textbf{(w/o 0h) $\downarrow$}} \\
\midrule
\multicolumn{6}{c}{\underline{\textit{Within-Speaker ABX} }} \\
Multi-Task-PT~[SSL] & 8.84 &	8.34 &	7.33 &	6.12 &	7.26 \\
~~~~~~~ + MAdaPT-Reptile & 7.97 &	7.10 &	6.61 &	5.79 &	6.50 \\
\rowcolor{gray!10}
~~~~~~~ + MAdaPT-FOBLO & 13.89 &	7.70 &	6.21 &	\textbf{5.29} &	6.40 \\
Multi-Task-PT~[SSL/SL] & \underline{7.57} &	6.84 &	6.20 &	5.65 &	6.23 \\
~~~~~~~ + MAdaPT-Reptile & \textbf{7.05} &	\underline{6.40} &	\underline{6.04} &	5.59 &	\underline{6.01} \\
\rowcolor{gray!10}
~~~~~~~ + MAdaPT-FOBLO & 7.73 &	\textbf{6.25} &	\textbf{5.99} &	\underline{5.56}	&	\textbf{5.93} \\
\midrule
\multicolumn{6}{c}{\underline{\textit{Across-Speaker ABX}}} \\
Multi-Task-PT~[SSL] & 10.48	& 9.77	& 8.18 &	6.80	&	8.25 \\
~~~~~~~ + MAdaPT-Reptile & 9.14 &	8.05 &	7.50 &	6.58	&	7.38 \\
\rowcolor{gray!10}
~~~~~~~ + MAdaPT-FOBLO & 16.28 &	8.61 &	6.78 &	\textbf{5.82}	&	7.07 \\
Multi-Task-PT~[SSL/SL] & \underline{8.23} &	7.40 &	6.50 &	6.06 &	6.65 \\
~~~~~~~ + MAdaPT-Reptile & \textbf{7.72} &	\underline{6.93} &	\underline{6.40} &	6.05 &	\underline{6.46}\\
\rowcolor{gray!10}
~~~~~~~ + MAdaPT-FOBLO & 8.40 &	\textbf{6.82} &	\textbf{6.37} &	\underline{5.96}	& \textbf{6.38} \\
\bottomrule
\end{tabular}

\end{table*}

\subsection{Detailed Results of Spoken Language Modeling}
\label{app:slm-details}
Detailed results for downstream spoken language modeling are provided under Table \ref{tab:detailed_slm}. As described in Experiment~\ref{experiment:slm}, we used sWUGGY, sBLIMP, and spoken tSC to estimate performance of the spoken language models. For all tasks, candidates are scored by length-normalized log-likelihood (log-likelihood divided by token count) for comparability across strings, and decisions are made by selecting the higher-scoring alternative.

We use SSL models finetuned on the English adaptation sets (0 hours to 100 hours) as encoders for the downstream SLM. A K-means model (with 256 units) is trained on the model embeddings to produce discrete tokens for language modeling. OPT-1.3B models \cite{zhang2022optopenpretrainedtransformer} are used as the SLMs, trained with fairseq2 \cite{balioglu2023fairseq2} and following the architectural decisions made by previous works \cite{hassid2023textually, Poli2025SpidR}. The full 60k hour dataset of Libri-Light \cite{9052942} is used for training. We train on 16 GPUs, with a context length of $2048$, and a batch of at most $40960$ tokens, for 150000 steps. The learning rate is set at $1e-3$ with a 1000-step warmup period and with a cosine annealing schedule. Remaining hyperparameters follow OPT-1.3B defaults. We select the checkpoint with the lowest validation loss.

\begin{table*}[ht]
    \centering
    \caption{\textbf{Detailed results of spoken language modeling metrics: sWUGGY, sBLIMP, and spoken tSC (in \%).} MAdaPT-FOBLO shows consistently superior performance across tasks. The best scores are in \textbf{bold} and the second best are \underline{underlined}.}
    \label{tab:detailed_slm}
    \renewcommand{\arraystretch}{1.1}
\setlength{\tabcolsep}{8pt}
\begin{tabular}{lcccccc}
\toprule
\textbf{Method} & \textbf{0h} & \textbf{10m} & \textbf{1h} & \textbf{10h} & \textbf{100h} & \makecell{\textbf{Avg.} \\\ \textbf{(w/o 0h) $\downarrow$}} \\
\midrule
\multicolumn{7}{c}{\textit{\underline{sWUGGY (in-vocab and out-of-vocab)}}} \\
In-Domain Mono-Task-PT                  &  \multicolumn{6}{c}{\textit{$6000$ hours training; Topline score is $64.51$.}}\\
\midrule
Multi-Task-PT~[SSL]                     
& 63.74 & 64.07 &
65.28 &
64.85 & 65.94 & 65.04 \\
~~~~~~~ + MAdaPT-Reptile          
& 64.17 & 63.99 & 64.53 & 64.54 & 64.88 & 64.49 \\
\rowcolor{gray!10}
~~~~~~~ + MAdaPT-FOBLO         & 54.82 & 64.42 & 64.60 & 66.30 & \textbf{67.46} & 65.70 \\
Multi-Task-PT~[SSL/SL]                      & 62.68 & 65.17 & 65.76 & 66.53 & 66.02 & 65.87 \\
~~~~~~~ + MAdaPT-Reptile         & \textbf{65.65} & 
\textbf{66.37} & \underline{66.92} & \textbf{67.79} & 67.25 & \textbf{67.08} \\
\rowcolor{gray!10}
~~~~~~~ + MAdaPT-FOBLO          & \underline{64.32} & \underline{66.04} & \textbf{66.99} & \underline{67.26} & \underline{67.26} & \underline{66.89} \\

\midrule
\multicolumn{7}{c}{\textit{\underline{sBLIMP}}} \\
In-Domain Mono-Task-PT                  &  \multicolumn{6}{c}{\textit{$6000$ hours training; Topline score is $56.94$.}}\\
\midrule
Multi-Task-PT~[SSL] & 55.73 &	56.07 &	57.12 &	56.46 &	57.94 &	56.90 \\
~~~~~~~ + MAdaPT-Reptile          & 56.23 &	56.88 &	56.75 &	57.54 & 56.84 & 57.00 \\
\rowcolor{gray!10}
~~~~~~~ + MAdaPT-FOBLO               & 52.93 &	\textbf{57.60} &	\underline{57.46} &	\underline{58.17} &	\textbf{58.11} &	\textbf{57.84} \\
Multi-Task-PT~[SSL/SL]                      & \underline{56.37} &	56.59 &	56.99 &	56.74 &	56.60 &	56.73 \\
~~~~~~~ + MAdaPT-Reptile         & \textbf{56.58} &	56.93 &	56.94 &	\textbf{58.47} &	\underline{58.01}	& 57.59 \\
\rowcolor{gray!10}
~~~~~~~ + MAdaPT-FOBLO          & 55.48 &	\underline{57.38} &	\textbf{57.99} &	57.54 &	57.87 & \underline{57.69} \\

\midrule
\multicolumn{7}{c}{\textit{\underline{Spoken tSC}}} \\
In-Domain Mono-Task-PT                  &  \multicolumn{6}{c}{\textit{$6000$ hours training; Topline score is $74.36$.}}\\
\midrule
Multi-Task-PT~[SSL]                     & 70.99 & 71.26 &	71.47 &	72.22 &	71.74 &	71.67 \\
~~~~~~~ + MAdaPT-Reptile          & 72.86 &	72.38 &	72.12 &	71.69 &	72.44 & 72.16  \\
\rowcolor{gray!10}
~~~~~~~ + MAdaPT-FOBLO               & 69.50 &	73.88 &	73.50 &	72.70 &	73.61 &	73.42 \\
Multi-Task-PT~[SSL/SL]                      & \underline{75.11} &	\textbf{74.25} &	\textbf{74.57} &	\underline{74.79} &	\underline{75.59} &	\textbf{74.80} \\
~~~~~~~ + MAdaPT-Reptile         & 74.95 &	73.02 &	\underline{74.20} &	74.04 &	\textbf{75.75} &	74.25 \\
\rowcolor{gray!10}
~~~~~~~ + MAdaPT-FOBLO          & \textbf{76.12} &	\underline{73.93} &	73.99 &	\textbf{74.84} &	75.43 &	\underline{74.55} \\
\bottomrule
\end{tabular}
\end{table*}

\subsection{Detailed Results on Phoneme Discovery}
\label{app:detailed_pd_benchmark}

DiscoPhon \cite{Poli2026Phoneme} is a benchmark specifically designed to investigate the abilities of speech representation models to encode phonemic information in a low resource setting. The benchmark includes 6 development languages (Swahili, Tamil, Thai, Ukrainian, Turkish, and German) and 6 test languages (French, English, Japanese, Mandarin, Wolof, and Basque) selected to span a diverse range of phonemic categories. Note that the development and test language sets in the benchmark differ from our previous experiments but are still disjoint from our training set. 

In the current experiment, we apply our previously tuned Multi-Task-PT, MAdaPT-Reptile, and MAdaPT-FOBLO models with SpidR as backbone. Our models are trained on 19 \texttt{VoxPopuli} languages \cite{wang-etal-2021-voxpopuli}. We compare our approaches to OoD HuBERT \cite{hsu2021hubert} and DinoSR \cite{liu2023dinosr} models trained on 20 \texttt{VoxPopuli} languages. Notably, supervised ASR encoders such as Whisper \citep{pmlr-v202-radford23a} exhibit poor phonemic discriminability despite strong recognition accuracy \citep{Poli2025SpidR}, and are therefore not included in our comparisons. For HuBERT and DinoSR, the best performing layer for ABX as determined by the development languages is used. For discrete unit metrics (PNMI and PER) a K-means is trained on HuBERT embeddings while the model codebooks are used for DinoSR and SpidR (with 256 codewords across all models). 

Test languages results are reported in Table~\ref{tab:detailed_pdb_test} and development languages results in Table~\ref{tab:detailed_pdb_dev}. As can be observed, on aggregate, our proposed MAdaPT-FOBLO achieves improved performance over alternate speech SSL models (HuBERT, DinoSR), with MAdaPT-Reptile also demonstrating strong performance.

\begin{table*}[ht]
    \centering
    \caption{\textbf{Detailed results (in \%) on DiscoPhon on the 6 test languages} (Mandarin Chinese, English, Basque, French, Japanese, Wolof). Average ABX for within- and across-speaker conditions is reported. MAdaPT-FOBLO and MAdaPT-Reptile outperform alternate speech SSL models (HuBERT, DinoSR). The best scores are in \textbf{bold} and the second best are \underline{underlined}.}
    \label{tab:detailed_pdb_test}
    \renewcommand{\arraystretch}{0.9}
\setlength{\tabcolsep}{10pt}
\begin{tabular}{lcccccc}
\toprule
\textbf{Method}  & \textbf{\makecell{Backbone \\ Model}} &  \textbf{0h} & \textbf{10m} & \textbf{1h} & \textbf{10h} & \makecell{\textbf{Avg.} \\ \textbf{(w/o 0h)}} \\
\midrule
\multicolumn{7}{c}{\textit{\underline{PER} $\downarrow$}} \\
Multi-Task-PT~[SSL] & HuBERT  & 127.02 & 114.12 & 101.03 & 98.02 & 104.39 \\
Multi-Task-PT~[SSL] & DinoSR & 86.32 &	78.40 &	70.14 &	69.79 &	72.78 \\
\midrule
Multi-Task-PT~[SSL] & SpidR  & 85.40 & 75.24 & 56.51 & 48.63 & 60.13 \\
~~~~~~~ + MAdaPT-Reptile       
& SpidR  & 153.77 & 66.00 & 60.03 & 54.42 & 60.15 \\
\rowcolor{gray!10}
~~~~~~~ + MAdaPT-FOBLO & SpidR  & 87.40 & 66.94 & 41.93 & \textbf{35.49} & 48.12 \\
Multi-Task-PT~[SSL/SL]   &  SpidR  & \textbf{50.82} & 40.94 & 38.77 & 37.99 & 39.23 \\
~~~~~~~ + MAdaPT-Reptile & SpidR  & 110.71 & \underline{40.77} & \textbf{36.88} & 37.05 & \underline{38.23} \\
\rowcolor{gray!10}
~~~~~~~ + MAdaPT-FOBLO  & SpidR  & \underline{51.06} & \textbf{39.32} & \underline{37.23} & \underline{36.58} & \textbf{37.71} \\

\midrule
\multicolumn{7}{c}{\textit{\underline{$R$-value} $\uparrow$}} \\
Multi-Task-PT~[SSL] & HuBERT  & 9.39 & 13.82 & 22.81 & 24.37 & 20.34\\
Multi-Task-PT~[SSL] & DinoSR & 38.93 &	41.35 &	45.38 &	43.96 &	43.56 \\
\midrule
Multi-Task-PT~[SSL] & SpidR  & 39.96 & 45.46 & 57.29 & 61.84 & 54.86 \\
~~~~~~~ + MAdaPT-Reptile       
& SpidR  & -7.66 & 52.87 & 55.61 & 58.50 & 55.66 \\
\rowcolor{gray!10}
~~~~~~~ + MAdaPT-FOBLO & SpidR  & 43.98 & 54.61 & 70.34 & 73.44 & 66.13 \\
Multi-Task-PT~[SSL/SL]   &  SpidR  & \textbf{67.27} & 71.21 & 72.03 & 72.78 & 72.01 \\
~~~~~~~ + MAdaPT-Reptile & SpidR  & 26.05 & \underline{72.37} & \textbf{75.28} & \textbf{75.01} & \textbf{74.22} \\
\rowcolor{gray!10}
~~~~~~~ + MAdaPT-FOBLO  & SpidR  & \underline{67.26} & \textbf{72.47} & \underline{73.50} & \underline{73.91} & \underline{73.29} \\

\midrule
\multicolumn{7}{c}{\textit{\underline{$F_1$} $\uparrow$}} \\
Multi-Task-PT~[SSL] & HuBERT  & 58.18 & 58.98 & 60.65 & 61.16 & 60.26 \\
Multi-Task-PT~[SSL] & DinoSR & 62.44 &	65.38 &	66.65 &	66.63 &	66.22 \\
\midrule
Multi-Task-PT~[SSL] & SpidR  & 63.88 & 65.46 & 69.69 & 71.10 & 68.75 \\
~~~~~~~ + MAdaPT-Reptile       
& SpidR  & 53.18 & 66.01 & 66.94 & 66.75 & 66.57 \\
\rowcolor{gray!10}
~~~~~~~ + MAdaPT-FOBLO & SpidR  & 29.55 & 69.54 & 76.06 & 77.05 & 74.22 \\
Multi-Task-PT~[SSL/SL]   &  SpidR  & \textbf{74.50} & 75.99 & 76.89 & \underline{77.52} & 76.80 \\
~~~~~~~ + MAdaPT-Reptile & SpidR  & 60.46 & \textbf{76.78} & \textbf{77.72} & \textbf{78.14} & \textbf{77.55} \\
\rowcolor{gray!10}
~~~~~~~ + MAdaPT-FOBLO  & SpidR  & \underline{74.24} & \underline{76.62} & \underline{77.06} & 77.51 & \underline{77.06} \\
\midrule
\multicolumn{7}{c}{\textit{\underline{PNMI} $\uparrow$}} \\
Multi-Task-PT~[SSL] & HuBERT  & 49.49 & 52.74 & 55.72 & 57.56 & 55.34 \\
Multi-Task-PT~[SSL] & DinoSR & 54.82 &	61.06 &	64.85 &	65.91 &	63.94 \\
\midrule
Multi-Task-PT~[SSL] & SpidR  & 58.25 & 61.89 & 67.16 & 69.34 & 66.13 \\
~~~~~~~ + MAdaPT-Reptile       
& SpidR  & 40.45 & 61.15 & 62.96 & 64.62 & 62.91 \\
\rowcolor{gray!10}
~~~~~~~ + MAdaPT-FOBLO & SpidR  & 10.36 & 62.03 & 68.70 & 70.78 & 67.17 \\
Multi-Task-PT~[SSL/SL]   &  SpidR  & \textbf{66.69} & \textbf{70.74} & \textbf{72.61} & \textbf{73.11} & \textbf{72.15} \\
~~~~~~~ + MAdaPT-Reptile & SpidR  & 48.90 & 67.86 & 68.96 & 69.42 & 68.75 \\
\rowcolor{gray!10}
~~~~~~~ + MAdaPT-FOBLO  & SpidR  & \underline{66.48} & \underline{70.12} & \underline{71.25} & \underline{71.64} & \underline{71.00} \\

\midrule
\multicolumn{7}{c}{\textit{\underline{ABX} $\downarrow$}} \\
Multi-Task-PT~[SSL] & HuBERT  & 7.79 & 7.63 & 6.77 & 5.49 & 6.63 \\
Multi-Task-PT~[SSL] & DinoSR & 8.48 &	7.73 &	7.20 &	6.64 &	7.19 \\
\midrule
Multi-Task-PT~[SSL] & SpidR  & 6.45 & 6.05 & 5.23 & 4.47 & 5.25 \\
~~~~~~~ + MAdaPT-Reptile       
& SpidR  & 5.76 & 5.25 & 4.91 & 4.30 & 4.82 \\
\rowcolor{gray!10}
~~~~~~~ + MAdaPT-FOBLO & SpidR  & 12.30 & 5.76 & 4.46 & \textbf{3.88} & 4.70 \\
Multi-Task-PT~[SSL/SL]   &  SpidR  & \underline{5.54} & 4.82 & 4.28 & 4.13 & 4.41 \\
~~~~~~~ + MAdaPT-Reptile & SpidR  & \textbf{5.07} & \underline{4.61} & \underline{4.28} & 4.11 & \underline{4.33} \\
\rowcolor{gray!10}
~~~~~~~ + MAdaPT-FOBLO  & SpidR  & 5.60 & \textbf{4.56} & \textbf{4.26} & \underline{4.10} & \textbf{4.31} \\
\bottomrule
\end{tabular}
\end{table*}

\begin{table*}[ht]
    \centering
    \caption{\textbf{Detailed results (in \%) on DiscoPhon on the 6 development languages} (German, Swahili, Tamil, Thai, Turkish, Ukrainian). Average ABX for within- and across-speaker conditions is reported. MAdaPT-FOBLO and MAdaPT-Reptile outperform alternate speech SSL models (HuBERT, DinoSR). The best scores are in \textbf{bold} and the second best are \underline{underlined}.}
    \label{tab:detailed_pdb_dev}
    \renewcommand{\arraystretch}{0.9}
\setlength{\tabcolsep}{10pt}
\begin{tabular}{lcccccc}
\toprule
\textbf{Method} & \textbf{\makecell{Backbone \\ Model}} & \textbf{0h} & \textbf{10m} & \textbf{1h} & \textbf{10h} & \makecell{\textbf{Avg.} \\ \textbf{(w/o 0h)}} \\
\midrule
\multicolumn{7}{c}{\textit{\underline{PER} $\downarrow$}} \\
Multi-Task-PT~[SSL] & HuBERT & 118.26 & 105.34 & 100.19 & 95.90 & 100.48 \\
Multi-Task-PT~[SSL] & DinoSR & 79.75 &	77.12 &	70.97 &	65.18 &	71.09 \\
\midrule
Multi-Task-PT~[SSL] & SpidR
 & 81.41 & 76.02 & 58.22 & 51.47 & 61.90 \\
~~~~~~~ + MAdaPT-Reptile & SpidR & 146.87 & 67.26 & 61.14 & 55.41 & 61.27 \\
\rowcolor{gray!10}
~~~~~~~ + MAdaPT-FOBLO & SpidR & 85.51 & 65.95 & 44.65 & \textbf{37.80} & 49.46 \\
Multi-Task-PT~[SSL/SL] & SpidR & \textbf{47.80} & 45.63 & 40.37 & 39.73 & 41.91 \\
~~~~~~~ + MAdaPT-Reptile & SpidR & 100.44 & \textbf{42.89} & \textbf{38.69} & \underline{38.67} & \textbf{40.08} \\
\rowcolor{gray!10}
~~~~~~~ + MAdaPT-FOBLO & SpidR & \underline{48.19} & \underline{43.83} & \underline{39.69} & 39.93 & \underline{41.15} \\

\midrule
\multicolumn{7}{c}{\textit{\underline{$R$-value} $\uparrow$}} \\
Multi-Task-PT~[SSL] & HuBERT & 19.63 & 24.23 & 28.52 & 30.48 & 27.74 \\
Multi-Task-PT~[SSL] & DinoSR & 47.38 &	46.73 &	48.35 &	51.37 &	48.82 \\
\midrule
Multi-Task-PT~[SSL] & SpidR
 & 46.18 & 48.69 & 59.44 & 62.64 & 56.92 \\
~~~~~~~ + MAdaPT-Reptile & SpidR & 0.05 & 54.58 & 58.62 & 59.00 & 57.40 \\
\rowcolor{gray!10}
~~~~~~~ + MAdaPT-FOBLO & SpidR & 42.29 & 58.50 & 72.03 & 73.65 & 68.06 \\
Multi-Task-PT~[SSL/SL] & SpidR & \textbf{72.03} & 71.17 & 74.05 & 74.51 & 73.24 \\
~~~~~~~ + MAdaPT-Reptile & SpidR & 37.86 & \textbf{73.39} & \textbf{75.86} & \textbf{75.87} & \textbf{75.04} \\
\rowcolor{gray!10}
~~~~~~~ + MAdaPT-FOBLO & SpidR & \underline{71.85} & \underline{72.30} & \underline{74.20} & \underline{74.51} & \underline{73.67} \\

\midrule
\multicolumn{7}{c}{\textit{\underline{$F_1$} $\uparrow$}} \\
Multi-Task-PT~[SSL] & HuBERT & 58.87 & 59.88 & 60.41 & 61.45 & 60.58 \\
Multi-Task-PT~[SSL] & DinoSR & 61.98 &	63.59 &	64.76 &	66.01	&	64.79 \\
\midrule
Multi-Task-PT~[SSL] & SpidR
 & 63.13 & 64.21 & 67.37 & 68.62 & 66.73 \\
~~~~~~~ + MAdaPT-Reptile & SpidR & 53.20 & 63.96 & 66.13 & 64.53 & 64.87 \\
\rowcolor{gray!10}
~~~~~~~ + MAdaPT-FOBLO & SpidR & 26.25 & 68.76 & 74.13 & 74.42 & 72.44 \\
Multi-Task-PT~[SSL/SL] & SpidR & \textbf{73.83} & \underline{74.25} & \textbf{75.77} & \underline{76.19} & \underline{75.40} \\
~~~~~~~ + MAdaPT-Reptile & SpidR & 60.42 & \textbf{74.62} & 75.51 & \textbf{76.39} & \textbf{75.51} \\
\rowcolor{gray!10}
~~~~~~~ + MAdaPT-FOBLO & SpidR & \underline{73.48} & 74.22 & \underline{75.62} & 76.14 & 75.33 \\

\midrule
\multicolumn{7}{c}{\textit{\underline{PNMI} $\uparrow$}} \\
Multi-Task-PT~[SSL] & HuBERT & 47.01 & 51.04 & 52.18 & 54.66 & 52.62 \\
Multi-Task-PT~[SSL] & DinoSR & 52.46  & 57.36 &	60.60 &	62.70	&	60.22 \\
\midrule
Multi-Task-PT~[SSL] & SpidR
 & 55.09 & 58.22 & 63.20 & 65.83 & 62.42 \\
~~~~~~~ + MAdaPT-Reptile & SpidR & 37.80 & 58.22 & 60.13 & 62.06 & 60.14 \\
\rowcolor{gray!10}
~~~~~~~ + MAdaPT-FOBLO & SpidR & 9.78 & 59.62 & 65.11 & 67.44 & 64.06 \\
Multi-Task-PT~[SSL/SL] & SpidR & \textbf{64.01} & \textbf{67.14} & \textbf{69.28} & \textbf{70.01} & \textbf{68.81} \\
~~~~~~~ + MAdaPT-Reptile & SpidR & 47.01 & 64.57 & 66.37 & 66.97 & 65.97 \\
\rowcolor{gray!10}
~~~~~~~ + MAdaPT-FOBLO & SpidR & \underline{63.75} & \underline{66.43} & \underline{68.19} & \underline{68.67} & \underline{67.76} \\

\midrule
\multicolumn{7}{c}{\textit{\underline{ABX} $\downarrow$}} \\
Multi-Task-PT~[SSL] & HuBERT & 10.23 & 9.49 & 9.08 & 7.68 & 8.75 \\
Multi-Task-PT~[SSL] & DinoSR & 11.44 &	10.27 &	9.94 &	9.02 &	9.74 \\
\midrule
Multi-Task-PT~[SSL] & SpidR
 & 8.79 & 8.26 & 7.12 & 6.02 & 7.13 \\
~~~~~~~ + MAdaPT-Reptile & SpidR & 7.80 & 6.95 & 6.48 & 5.75 & 6.39 \\
\rowcolor{gray!10}
~~~~~~~ + MAdaPT-FOBLO & SpidR & 14.34 & 7.46 & 5.98 & \textbf{5.15} & 6.20 \\
Multi-Task-PT~[SSL/SL] & SpidR & \underline{7.19} & 6.54 & 5.84 & 5.40 & 5.93 \\
~~~~~~~ + MAdaPT-Reptile & SpidR & \textbf{6.73} & \underline{6.11} & \underline{5.69} & 5.34 & \underline{5.72} \\
\rowcolor{gray!10}
~~~~~~~ + MAdaPT-FOBLO & SpidR & 7.30 & \textbf{6.01} & \textbf{5.67} & \underline{5.31} & \textbf{5.66} \\
\bottomrule
\end{tabular}
\end{table*}

\section{Ablation Studies}

\subsection{Impact of Active Forgetting}
\label{app:active_forgetting}
To investigate the impact of active forgetting in our approach we conduct ablation studies by removing the active forgetting mechanism from the inner loop on the 5 development and 3 test languages. As shown in Table~\ref{tab:active_forgetting_5dev} and Table~\ref{tab:active_forgetting_3test}, incorporating active forgetting consistently outperforms the variant without this mechanism. This demonstrates that resetting the prediction heads and codebooks helps the model alleviate overfitting to previous episodes, thereby improving overall performance.

\begin{table*}[ht]
    \centering
    \caption{\textbf{Impact of active forgetting on \textcolor{blue}{$5$ DEVELOPMENT} languages}. \cmark~ and \xmark~ denote whether we deploy the active forgetting mechanism in the inner loop or not, respectively. Broadly, active forgetting improves adaptation performance, preventing overfitting to training languages. The best scores are in \textbf{bold} and second best are \underline{underlined}.}
    \label{tab:active_forgetting_5dev}
    \renewcommand{\arraystretch}{1.2}
\setlength{\tabcolsep}{4pt}
\begin{tabular}{lccccc>{\columncolor{gray!10}}c}
\toprule
Method & {Active Forgetting} & \textbf{0h} & \textbf{10m} & \textbf{1h} & \textbf{10h} & \makecell{\textbf{Avg.} \\\ \textbf{(w/o 0h) $\downarrow$}} \\
\midrule
\multicolumn{7}{c}{\underline{\textit{Within-Speaker ABX}}} \\
\multirow{2}{5cm}{Multi-Task-PT~[SSL]~~ + MAdaPT-FOBLO} & \xmark  & 10.68 &	\underline{6.45} &	6.75 &	6.63 &	6.61 \\
& \cmark & 13.89 &	7.70 &	6.21 &	\textbf{5.29} &	6.40 \\
\midrule
\multirow{2}{5cm}{Multi-Task-PT~[SSL/SL]~~ + MAdaPT-FOBLO} & \xmark  & \textbf{7.45} &	6.74 &	\underline{6.10} &	5.57 &	\underline{6.14}\\
& \cmark  & \underline{7.73} &	\textbf{6.25} &	\textbf{5.99} &	\underline{5.56} &	\textbf{5.93} \\

\midrule
\multicolumn{7}{c}{\underline{\textit{Across-Speaker ABX}}} \\
\multirow{2}{5cm}{Multi-Task-PT~[SSL]~~ + MAdaPT-FOBLO} & \xmark  & 12.38 &	\underline{7.13} &	7.09 &	6.78 &	7.00 \\
& \cmark & 16.28 &	8.61 &	6.78 &	\textbf{5.82} &	7.07 \\
\midrule
\multirow{2}{5cm}{Multi-Task-PT~[SSL/SL]~~ + MAdaPT-FOBLO} & \xmark  & \textbf{8.12} &	7.24 &	\underline{6.46} &	6.01 &	\underline{6.57}\\
& \cmark  & \underline{8.40} &	\textbf{6.82} &	\textbf{6.37} &	\underline{5.96}  &	\textbf{6.38}\\
\bottomrule
\end{tabular}
\end{table*}

\begin{table*}[ht]
    \centering
    \caption{\textbf{Impact of active forgetting on \textcolor{blue}{$3$ TEST} languages}. \cmark~ and \xmark~ denote whether we deploy the active forgetting mechanism in the inner loop or not, respectively. Similar to results in development languages, active forgetting here improves adaptation performance, preventing overfitting to training languages. The best scores are in \textbf{bold} and second best are \underline{underlined}.}
    \label{tab:active_forgetting_3test}
    \renewcommand{\arraystretch}{1.2}
\setlength{\tabcolsep}{3pt}
\begin{tabular}{lcccccc>{\columncolor{gray!10}}c}
\toprule
Method & {Active Forgetting} & \textbf{0h} & \textbf{10m} & \textbf{1h} & \textbf{10h} & \textbf{100h} & \makecell{\textbf{Avg.} \\\ \textbf{(w/o 0h) $\downarrow$}} \\
\midrule
\multicolumn{8}{c}{\underline{\textit{Within-Speaker ABX}}} \\
In-Domain Mono-Task-PT &{N.A.} & \multicolumn{6}{c}{\textit{$6000$ hours training; Topline score is $4.10$.}}\\
\midrule
\multirow{2}{5cm}{Multi-Task-PT~[SSL]~~ + MAdaPT-FOBLO} & \xmark  & 21.89 & 4.40 & 4.54 & 4.37 & 4.20 & 4.38 \\ 
& \cmark  & 10.05 & 4.51 & 4.05 & 3.78 & \underline{3.69} & 4.01 \\ 
\midrule
\multirow{2}{5cm}{Multi-Task-PT~[SSL/SL]~~ + MAdaPT-FOBLO} & \xmark  & \textbf{3.99} & \textbf{4.02} & \underline{3.87} & \underline{3.71} & \textbf{3.67} & \underline{3.82} \\ 
& \cmark & \underline{4.00} & \underline{4.07} & \textbf{3.84} & \textbf{3.62} & 3.70 & \textbf{3.80} \\

\midrule
\multicolumn{8}{c}{\underline{\textit{Across-Speaker ABX}}} \\
In-Domain Mono-Task-PT &{N.A.} & \multicolumn{6}{c}{\textit{$6000$ hours training; Topline score is $5.47$.}}\\
\midrule
\multirow{2}{5cm}{Multi-Task-PT~[SSL]~~ + MAdaPT-FOBLO} & \xmark  & 29.11 & 5.62 & 5.77 & 5.56 & 5.33 & 5.57 \\ 
& \cmark  & 15.12 & 5.96 & 5.26 & 4.92 & 4.83 & 5.24 \\  
\midrule
\multirow{2}{5cm}{Multi-Task-PT~[SSL/SL]~~ + MAdaPT-FOBLO} & \xmark  & \underline{5.29} & \underline{5.32} & \underline{5.01} & \underline{4.84} & \underline{4.80} & \underline{4.99} \\ 
& \cmark & \textbf{5.28} & \textbf{5.23} & \textbf{4.96} & \textbf{4.76} & \textbf{4.77} & \textbf{4.93} \\ 
\bottomrule
\end{tabular}
\end{table*}

\subsection{Impact of Meta-Initialization}
\label{app:initialization}
To explore the influence of meta-initialization, we meta-train our model from three types of initialization. Multi-Task-PT [SSL] and Multi-Task-PT [SSL/SL] have been introduced in the main manuscript, both obtained via multi-task pre-training. Here we attempt random initialization, wherein the backbone is initialized by random sampling from the default parameter distribution. Table~\ref{tab:meta_initialization_ABX_5dev} and Table~\ref{tab:meta_initialization_ABX_3test} present ablation studies with different meta-initializations on 5 development and 3 test languages, respectively.

We find that random meta-initialization does not work for meta-training. Without a meaningful starting point, meta-training may fail to converge or require significantly more data and iterations to achieve competitive performance for self-supervised speech models. Thus, the success of meta-learning for speech representation learning is tightly coupled with the quality and relevance of the initial representations encoded in the backbone.

\begin{table*}[ht]
    \centering
    \caption{\textbf{Impact of meta-initialization on \textcolor{blue}{$5$ DEVELOPMENT} languages}. Random initialization produces unstable model training. The best scores are in \textbf{bold} and second best are \underline{underlined}.}
    \label{tab:meta_initialization_ABX_5dev}
    \begin{tabular}{l|l|cccc>{\columncolor{gray!10}}c}
\toprule
Method & {Meta-Initialization} & \textbf{0h} & \textbf{10m} & \textbf{1h} & \textbf{10h} & \makecell{\textbf{Avg.} \\\ \textbf{(w/o 0h) $\downarrow$}} \\
\midrule
\multicolumn{7}{c}{\underline{\textit{Within-Speaker ABX}}} \\
\multirow{3}{*}{ MAdaPT-FOBLO}  & Random  & 35.83 &	31.22 &	35.73 &	37.37 &	34.77 \\
& Multi-Task-PT~[SSL] & \underline{13.89} &	\underline{7.70} &	\underline{6.21} &	\textbf{5.29} &	\underline{6.40} \\
& Multi-Task-PT~[SSL/SL] & \textbf{7.73} &	\textbf{6.25} &	\textbf{5.99} &	\underline{5.56} &	\textbf{5.93} \\
\midrule
\multicolumn{7}{c}{\underline{\textit{Across-Speaker ABX}}} \\
\multirow{3}{*}{MAdaPT-FOBLO}  & Random & 8.12 &	7.24 &	6.46 &	6.01 &	6.57 \\
& Multi-Task-PT~[SSL] & \underline{16.28} &	\underline{8.61} &	\underline{6.78} &	\textbf{5.82} &	\underline{7.07} \\
& Multi-Task-PT~[SSL/SL] & \textbf{8.40} &	\textbf{6.82} &	\textbf{6.37} &	\underline{5.96} &	\textbf{6.38} \\
\bottomrule
\end{tabular}
\end{table*}

\begin{table*}[ht]
    \centering
    \caption{\textbf{Impact of meta-initialization on \textcolor{blue}{$3$ TEST} languages}. Random initialization produces unstable model training. The best scores are in \textbf{bold} and second best are \underline{underlined}.}
    \label{tab:meta_initialization_ABX_3test}
    \begin{adjustbox}{width=0.9\linewidth}
    \begin{tabular}{l|c|ccccc>{\columncolor{gray!10}}c}
\toprule
Method & {Meta-Initialization} & \textbf{0h} & \textbf{10m} & \textbf{1h} & \textbf{10h}  & \textbf{100h}  & \makecell{\textbf{Avg.} \\\ \textbf{(w/o 0h) $\downarrow$}} \\
\midrule
\multicolumn{8}{c}{\underline{\textit{Within-Speaker ABX}}} \\
\makecell{In-Domain \\\ Mono-Task-PT} & {N.A.} & \multicolumn{6}{c}{\textit{$6000$ hours training; Topline score is $4.10$.}} \\
\midrule
\multirow{3}{*}{ MAdaPT-FOBLO}  & Random  & 32.68 & 25.12 & 24.61 & 23.75 & 24.52 & 24.50 \\
&Multi-Task-PT~[SSL] & \underline{10.05} & \underline{4.51} & \underline{4.05} & \underline{3.78} & \textbf{3.69} & \underline{4.01} \\
& Multi-Task-PT~[SSL/SL] & \textbf{4.00} & \textbf{4.07} & \textbf{3.84} & \textbf{3.62} & \underline{3.70} & \textbf{3.80} \\
\midrule
\multicolumn{8}{c}{\underline{\textit{Across-Speaker ABX}}} \\
\makecell{In-Domain \\\ Mono-Task-PT} & {N.A.} & \multicolumn{6}{c}{\textit{$6000$ hours training; Topline score is $5.47$.}} \\
\midrule
\multirow{3}{*}{ MAdaPT-FOBLO}  & Random & 38.75 & 33.25 & 32.81 & 32.31 & 32.78 & 32.79 \\
&Multi-Task-PT~[SSL]  & \underline{15.12} & \underline{5.96} & \underline{5.26} & \underline{4.92} & \underline{4.83} & \underline{5.24} \\
& Multi-Task-PT~[SSL/SL] & \textbf{5.28} & \textbf{5.23} & \textbf{4.96} & \textbf{4.76} & \textbf{4.77} & \textbf{4.93} \\
\bottomrule
\end{tabular}
    \end{adjustbox}
\end{table*}

\subsection{Analysis of Meta-Learning Rate}
To systematically investigate the impact of the meta-learning rate $\beta$ in our approach, we conduct a series of experiments with SpidR-Adapt, utilizing interleaved supervised meta-initialization and varying $\beta$ across $0.001$, $0.01$, $0.1$ and $1$. 
Table~\ref{tab:hyper-beta_ABX_all} presents the results, with the left and right sub-tables corresponding to the 5 development and 3 test language sets, respectively. 
Our analysis reveals a clear trend on the development set as $\beta$ increases: ABX scores initially become lower, reaching its peak at $\beta=0.01$ before declining at higher values. This suggests that a moderate meta-learning rate strikes the best balance between adaptation and stability, while excessively high rates may lead to suboptimal generalization.

To ensure robust hyperparameter selection and prevent overfitting to the 3 test languages, we rely on the 5 development results to identify the optimal $\beta$. Consequently, all reported results in the paper are based on $\beta=0.01$, which consistently yields the strongest performance across our evaluation.

\begin{table*}[ht]
    \centering
    \caption{\textbf{Impact of meta-learning rate $\beta$ on \textcolor{blue}{$5$ DEVELOPMENT} and \textcolor{blue}{$3$ TEST} languages}. Best performing $\beta$ is $0.01$ for MAdaPT-FOBLO [SSL/SL] on development languages and is retained for test language inference. The best scores are in \textbf{bold} and second best are \underline{underlined}.}
    \label{tab:hyper-beta_ABX_all}
    \begin{adjustbox}{width=\linewidth}
    \normalsize
\begin{tabular}{lcccc>{\columncolor{gray!10}}c}
\toprule
$\bm \beta$ & \textbf{0h} & \textbf{10m} & \textbf{1h} & \textbf{10h} & \makecell{\textbf{Avg.} \\\ \textbf{(w/o 0h) $\downarrow$}} \\
\midrule
\multicolumn{6}{c}{\underline{\textit{Within-Speaker ABX} }} \\
$0.001$ & \textbf{7.59} &	6.33 &	6.06 &	5.59 &	6.00 \\
$0.01$  & \underline{7.73} &	\textbf{6.25} &	\underline{5.99} &	\textbf{5.56} &	\textbf{5.93} \\
$0.1$   & 11.63 &	\underline{6.31} &	\textbf{5.97} &	5.61 &	\underline{5.96} \\
$1$   &  8.23 &	6.63 &	6.09 &	\underline{5.58} &	6.10 \\
\midrule
\multicolumn{6}{c}{\underline{\textit{Across-Speaker ABX} }} \\
$0.001$ & \textbf{8.22} &	6.92 &	6.42 &	6.05 &	6.46 \\
$0.01$  & \underline{8.42} &	\textbf{6.82} &	\textbf{6.38} &	\textbf{5.96} &	\textbf{6.39} \\
$0.1$ &  12.66 &	\underline{6.84} &	\underline{6.41} &	\underline{6.04} &	\underline{6.43} \\
$1$  & 9.10 &	7.15 &	6.51 &	6.05 &	6.57 \\
\bottomrule
\end{tabular}

~~
\begin{tabular}{lccccc>{\columncolor{gray!10}}c}
\toprule
$\bm \beta$ & \textbf{0h} & \textbf{10m} & \textbf{1h} & \textbf{10h} & \textbf{100h} & \makecell{\textbf{Avg.} \\\ \textbf{(w/o 0h) $\downarrow$}} \\
\midrule
\multicolumn{7}{c}{\underline{\textit{Within-Speaker ABX} }} \\
$0.001$ & \underline{4.10} & \underline{4.02} & 3.87 & 3.66 & 3.69 & 3.81 \\
$0.01$  & \textbf{4.00}    & 4.07 & 3.84 & 3.62 & 3.70 & \underline{3.80} \\
$0.1$   & 6.20             & \textbf{3.91} & \underline{3.81} & \underline{3.57} & \underline{3.64} & \textbf{3.73} \\
$1$     & 5.89             & 4.04 & \textbf{3.72} & \textbf{3.53} & \textbf{3.62} & \textbf{3.73} \\
\midrule
\multicolumn{7}{c}{\underline{\textit{Across-Speaker ABX} }} \\
$0.001$ & \underline{5.39} & 5.25 & 4.97 & 4.79 & 4.77 & 4.95 \\
$0.01$  & \textbf{5.28}    & \underline{5.23} & \underline{4.96} & \underline{4.76} & 4.77 & 4.93 \\
$0.1$   & 8.56             & \textbf{5.16} & 5.00 & \underline{4.76} & \underline{4.75} & \underline{4.92} \\
$1$    & 7.87             & 5.29 & \textbf{4.86} & \textbf{4.68} & \textbf{4.72} & \textbf{4.89} \\
\bottomrule
\end{tabular}
    \end{adjustbox}
\end{table*}

\subsection{Layer-wise Analysis on the Model’s Discriminability.}
\begin{figure*}[ht]
\centering \includegraphics[width=\linewidth]{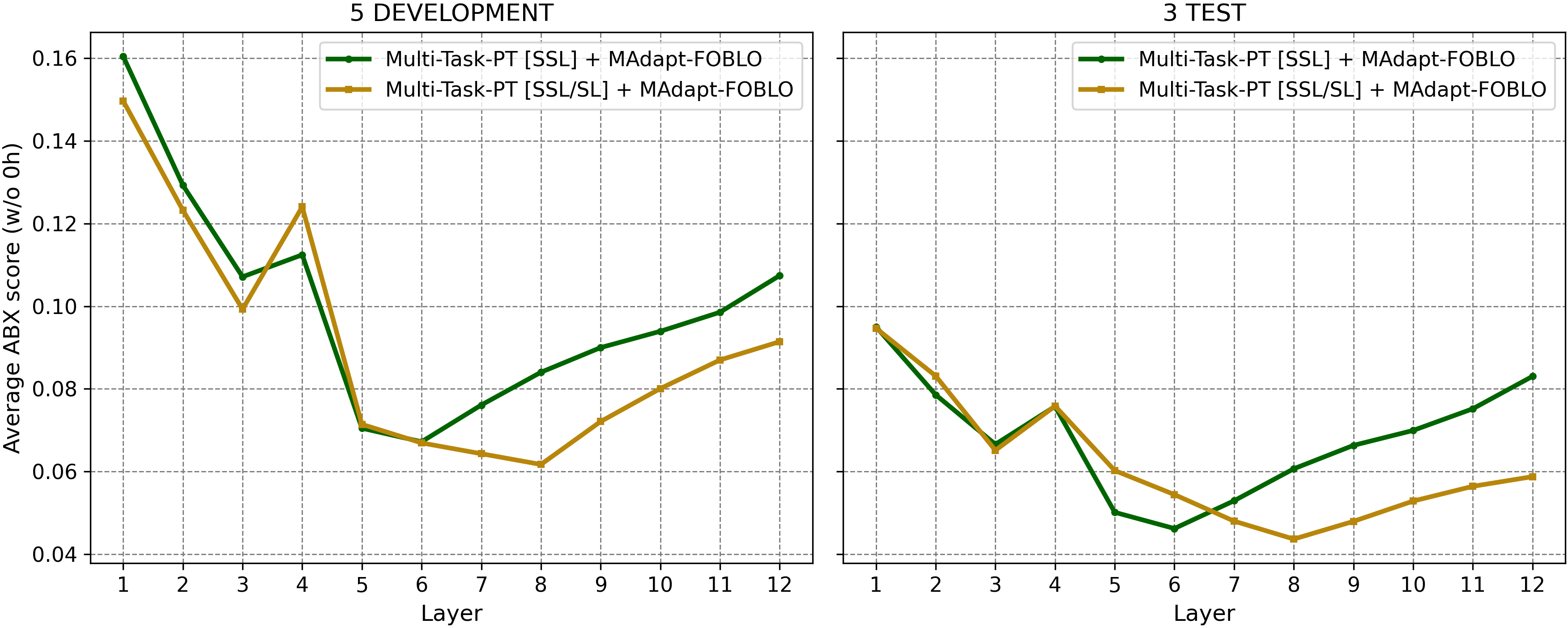}
\caption{\textbf{Layer-wise analysis on the model's discriminability over phonemes.} We present the ABX scores averaged over the target languages, and across the two within- and across-speaker conditions for two evaluation sets: (a) $5$ development  and (b) $3$ test language sets. We report results for our proposed MAdaPT-FOBLO method with two types of meta-initialization, Multi-Task-PT[SSL] and Multi-Task-PT[SSL/SL]. The optimal layer for ABX performance remains consistent across both ABX conditions, but varies depending on the meta-initialization. Specifically, the optimal layer is $6$ for Multi-Task-PT[SSL] initialization and $8$ for Multi-Task-PT[SSL/SL] initialization.}
\label{fig:layer_wise}
\end{figure*}

To investigate how layer-specific embeddings affect the model’s ability to discriminate between phonemes, we present ABX scores for each student layer in Figure~\ref{fig:layer_wise}. The scores are averaged across (a) 5 development or (b) 3 test languages and averaged across 10 minute to 100 hour adaptation data scales. Our analysis reveals distinct trends for different meta-initialization strategies applied to MAdaPT-FOBLO: 1) with Multi-Task-PT [SSL], 
the phone discriminability improves with increasing layer depth, peaking at layer $6$. Beyond this point, performance declines, suggesting that intermediate layers capture the most relevant phonetic representations, while deeper layers may become overly specialized or abstracted for the ABX task.
2) with Multi-Task-PT [SSL/SL], the optimal performance is observed at layer $8$. 

These results suggest that the best performing layer is consistent with the layer at which supervision is applied during the outer loop of FOBLO. For SSL meta-initialization, the a supervision head is attached to the 6th encoder layer for outer loop supervision while for SSL/SL meta-initialization, it is attached to the 8th layer (see Appendix~\ref{app:training_details} for more details here). 

\section{Extended Ablation Studies}
\label{app:extended}
\subsection{Impact of Interleaved Steps in the Multi-task Pretraining}
\begin{table*}[ht]
\centering
\caption{\textbf{Impact of interleaved steps in the multi-task pretraining on \textcolor{blue}{$5$ DEVELOPMENT} languages}. Overall performance is stable across higher frequency interleaving with smaller steps \{2, 10\} and degrades at steps larger than 50, revealing a trade-off between zero-shot transfer and few-shot adaptability.}
\label{tab: extended_ablation_interleavedSL}

\begin{tabular}{ccccccc}
\toprule
\makecell{\textbf{Interleaved Steps in} \\\ \textbf{Multi-Task-PT [SSL/SL] }} & \textbf{0h} & \textbf{10m} & \textbf{1h} & \textbf{10h} & \makecell{\textbf{Avg.} \\ \textbf{(w/o 0h) $\downarrow$}} \\
\midrule
\multicolumn{6}{c}{\underline{\textit{Within-Speaker ABX}}} \\
$2$ & 9.49 & 6.66 & 6.01 & 5.56 & 6.08 \\
$10$ & 7.57	& 6.84 & 6.20 & 5.65 & 6.23 \\
$50$ & 7.13 & 6.65 & 6.36 & 6.14 & 6.38 \\
$100$ & 7.01 & 6.81 & 6.41 & 6.14 & 6.45 \\
\midrule
\multicolumn{6}{c}{\underline{\textit{Across-Speaker ABX}}} \\
$2$ & 10.17 & 7.25 & 6.47 & 5.93 & 6.55 \\
$10$ & 8.23	& 7.40 & 6.50 & 6.06 & 6.65\\
$50$ & 7.88 & 7.30 & 6.83 & 6.52 & 6.88 \\
$100$ & 7.71 & 7.42 & 6.92 & 6.70 & 7.01 \\
\bottomrule
\end{tabular}
\end{table*}

We conduct an ablation over the number of interleaved supervision steps $\in \{2, 10, 50, 100\}$ (Table~\ref{tab: extended_ablation_interleavedSL}). We observe a trade-off between zero-shot and few-shot performance: more frequent interleaving improves zero-shot transfer by injecting stronger supervised structure, but degrades few-shot adaptability. Crucially, few-shot performance is roughly stable across steps $\{2, 10\}$, and only degrades noticeably at steps $\geq$ 50. We select steps = 10 as it offers the best balance: competitive few-shot performance with substantially improved zero-shot transfer over steps = 2.

\subsection{Analysis on Supervised Steps \texorpdfstring{$N$}{N} during meta-training}

\begin{table*}[ht]
\centering
\caption{\textbf{Impact of supervised steps during meta-training on \textcolor{blue}{$5$ DEVELOPMENT} languages}. Multi-Task-PT [SSL/SL] makes the meta-model more robust to variation in the number of supervised steps compared to Multi-Task-PT [SSL].}
\label{tab: extended_ablation_MandN}
\begin{tabular}{llccccc}
\toprule
\textbf{Method} & \textbf{Supervised Steps N} & \textbf{0h} & \textbf{10m} & \textbf{1h} & \textbf{10h} & \makecell{\textbf{Avg.} \\ \textbf{(w/o 0h) $\downarrow$}} \\
\midrule
\multicolumn{7}{c}{\underline{\textit{Within-Speaker ABX}}} \\
 & 0 & \multicolumn{5}{c}{\textit{codebook collapse}} \\
 Multi-Task-PT [SSL] + & 40 & 23.16 & 7.81 & 6.83 & 5.62 & 6.76 \\
MAdaPT-FOBLO & 200 & 13.89 & 7.70 &	6.21 & 5.29 & 6.40\\
 & 1000 & 8.02 & 6.62 & 5.80 & 5.16 & 5.86 \\
\cmidrule{1-7}
 & 0 & 7.62 & 6.53 & 6.16 & 5.64 & 6.11 \\
Multi-Task-PT [SSL/SL] + & 40 & 7.68 & 6.23 & 6.10 & 5.62 & 5.98 \\
MAdaPT-FOBLO & 200 & 7.73 & 6.25 & 5.99 & 5.56	& 5.93 \\
 & 1000 & 7.89 & 6.31 & 5.99 & 5.54 & 5.95 \\
\midrule
\multicolumn{7}{c}{\underline{\textit{Across-Speaker ABX}}} \\
& 0 & \multicolumn{5}{c}{\textit{codebook collapse}} \\
Multi-Task-PT [SSL] + & 40 & 27.31 & 8.78 & 7.45 & 6.36 & 7.53 \\
MAdaPT-FOBLO & 200 & 16.28 & 8.61 &	6.78 & 5.82	& 7.07 \\
 & 1000 & 9.28 & 7.42 & 6.45 & 5.60 & 6.49 \\
\cmidrule{1-7}
& 0 & 8.35 & 6.98 & 6.48 & 6.03 & 6.50 \\
Multi-Task-PT [SSL/SL] + & 40 & 8.34 & 6.77 & 6.36 & 5.96 & 6.36 \\
MAdaPT-FOBLO & 200 & 8.40 & 6.82 & 6.37 & 5.96	& 6.38 \\
 & 1000 & 8.51 & 6.77 & 6.28 & 5.91 & 6.32 \\
\bottomrule
\end{tabular}
\end{table*}

To investigate the impact of supervised steps during meta-training, we systematically vary $N \in {0, 40, 200, 1000}$ under two pre-training regimes: Multi-Task-PT [SSL/SL] and Multi-Task-PT [SSL]. As shown in Table~\ref{tab: extended_ablation_MandN}, with SSL/SL pre-training, increasing $N$ improves few-shot performance while slightly degrading zero-shot transfer, revealing a trade-off between initialization generality and adaptation depth. With SSL-only pre-training, $N=0$ leads to codebook collapse, confirming that supervised inner loop steps are essential for producing informative meta-gradients. Note that we use the same hyperparameter configuration ($N=200$) across all model variants, which may not be optimal for each setting; we leave this exploration to future work.
\end{document}